\documentclass[a4paper,12pt,times,numbered,print,index,oneside,custommargin]{Classes/PhDThesisPSnPDF}

\input{Preamble/preamble}

\title{A Tensor Network Implementation of Multi Agent Reinforcement Learning}

\author{Sunny Howard}

\dept{Physics and Astronomy}

\university{University of Nottingham}
\crest{\includegraphics[width=0.4\textwidth]{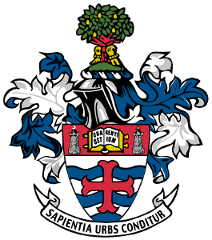}}

    \supervisor{Dr Edward Gillman}

\supervisorlinewidth{0.33\textwidth}

\advisor{Dr Dominic Rose}
     
\advisorlinewidth{0.28\textwidth}

\degreetitle{MSc in Machine Learning in Science}

\degreedate{August 2020} 

\subject{LaTeX} \keywords{{LaTeX} {Master Of Advanced Study} {Physics} {University of
Nottingham}}

\begin{document}

\frontmatter

\maketitle

\begin{abstract}
    Recently it has been shown that tensor networks (TNs) have the ability to represent the expected return of a single-agent finite Markov decision process (FMDP). The TN represents a distribution model, where all possible trajectories are considered. When extending these ideas to a multi-agent setting, distribution models suffer from the curse of dimensionality: the exponential relation between the number of possible trajectories and the number of agents.  The key advantage of using TNs in this setting is that there exists a large number of established optimisation and decomposition techniques that are specific to TNs, that one can apply to ensure the most efficient representation is found. In this report, these methods are used to form a TN that represents the expected return of a multi-agent reinforcement learning (MARL) task. This model is then applied to a 2 agent random walker example, where it was shown that the policy is correctly optimised using a DMRG technique. Finally, I demonstrate the use of an exact decomposition technique, reducing the number of elements in the tensors by 97.5\%, without experiencing any loss of information. 
\end{abstract}

\tableofcontents

\printnomenclature

\mainmatter

\chapter{Introduction}  

\ifpdf
    \graphicspath{{Chapter1/Figs/Raster/}{Chapter1/Figs/PDF/}{Chapter1/Figs/}}
\else
    \graphicspath{{Chapter1/Figs/Vector/}{Chapter1/Figs/}}
\fi

Since their invention, tensor networks (TNs) have been utilised in many areas of science. Although predominantly in the context of quantum physics, to calculate the ground state of a quantum many body systems \cite{tensorquant,tensorquant2,tensorquant3}, these concepts have also been employed in neuroscience \cite{braintensor} and in the solving of partial differential equations \cite{tensormath}. Recently, applications have spread to the field of computer science, and more specifically machine learning (ML) \cite{tnml1,tnml2,tnml3, tnml4,tnml5,TensorDeepLearning,tensordeeplearning3,tensordeeplearning4,tensordeeplearning5}. An indication of TNs functionally within ML is shown in the fact that Google's AI team recently collaborated in the release of an open-source library to make TN methods more accessible \cite{googletn}.  

\section{Tensors and Tensor Networks}\label{section:contraction}
\begin{figure}[h]
     \centering
     \begin{subfigure}[b]{0.3\textwidth}
         \centering
         \includegraphics[width=0.7\textwidth]{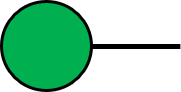}
         \caption{\centering Rank 1 Tensor, displayed by a circle}
         \label{fig:y equals x}
     \end{subfigure}
     \hfill
     \begin{subfigure}[b]{0.3\textwidth}
         \centering
         \includegraphics[width=\textwidth]{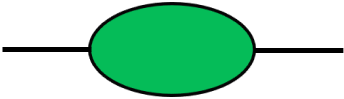}
         \caption{\centering Rank 2 Tensor, displayed by an oval}
         \label{fig:three sin x}
     \end{subfigure}
     \hfill
     \begin{subfigure}[b]{0.3\textwidth}
         \centering
         \includegraphics[width=0.7\textwidth]{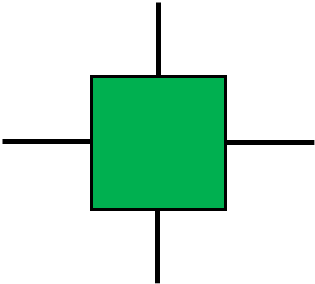}
         \caption{\centering Rank 4 Tensor, displayed by a square}
         \label{fig:rank4}
     \end{subfigure}
        \caption{The classical graphical representations of different ranked tensors. The number of lines coming off the green shape indicates how many indices the tensor has (and therefore its rank). In this report, these shapes will only be followed loosely. }
        \label{fig:definetensor}
\end{figure}

A tensor is a multidimensional array. The number of vectors that one must contract with the tensor in order to obtain a scalar is described by the rank. A rank 0 tensor is an object that requires no index, a scalar. A rank 1 tensor requires one index, which is a vector. The physical meaning of the rank, is that it describes the number of components within vectorspace that a relationship is being found between \cite{tensordefinition}. For example, a rank 2 tensor (a matrix) describes the relation between two vectors. Often when working with tensor networks, one will be using many tensors and therefore it is important to have a succinct way of displaying them. In 1971, Penrose graphical notation was invented for exactly this reason \cite{penrosegraphical}.  An example of how tensors are displayed graphically is shown in figure \ref{fig:definetensor}.

In this report the shapes shown in figure \ref{fig:definetensor} will only be followed loosely. This is because there will be tensors that have similar meanings, and it is clearer to mark these with a consistent shape, even if their ranks are not the same. To tell the rank, one must only look at the number of lines protruding from the shape. Further to this, sometimes the lines will be explicitly marked by the indices that they represent, and this is simply for clarity (there is no mathematical difference between the lines being marked and not marked).

\begin{figure}[h]
    \centering
    \includegraphics[width = 8cm]{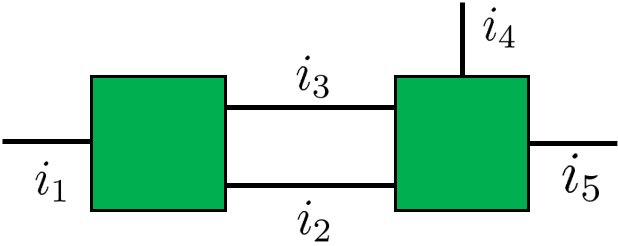}
    \caption{A figure showing the contraction of a rank 3 tensor with a rank 4 tensor. The indices are labelled to make it clear what the shared indices are, which in this case are $i_{2}$ and $i_{3}$.}
    \label{fig:tensorcontraction}
\end{figure}

A tensor network is defined to be a countable collection of tensors, connected by contractions \cite{TensorDefinition1}. It is therefore logical to also define the term contraction: a mathematical operation that sums over all possible values of repeated indices in two tensors \cite{contraction}. Graphically this is represented as the lines of different tensors connecting, as seen in figure \ref{fig:tensorcontraction}.

In a TN, a high rank tensor can be represented by the contractions of many lower rank tensors. This is useful because high rank tensors take up large quantities of memory to store and to operate on, due to the curse of dimensionality (the exponential relationship between tensor rank and number of elements) \cite{curseofdimensionality}. In fact, by decomposing them into lower rank tensors, calculations can become exponentially less computationally costly. 


\section{Tensor Networks In Reinforcement Learning}
Whilst there has been much work within both supervised and unsupervised machine learning, TNs have only been applied to reinforcement learning (RL) very recently \cite{edpaper}. For the most part, reinforcement learning models can be sorted into two categories: sample models and distribution models \cite{suttonbarto}, with the latter being what TNs can represent. Sample models act by sampling trajectories from a true environment, and then optimising a policy with respect to this information in order to maximise the agent's return. Distribution models consider all possible trajectories of an agent, along with their respective probabilities. Optimising the policy in a distribution model maximises a slightly different quantity, the expected return \cite{modelvsmodelbased}. Sample models can only act on information generated in previous trajectories, where as distribution models act on all possible trajectories, which in some sense indicates that they are superior. However, as every possible trajectory must be considered, they are also far more computationally expensive. Furthermore, in the general case the number of possible trajectories scales exponentially with both the number of timesteps, $T$, and the number of agents. If one were to store this information about possible trajectories in one tensor, the number of elements would also have an exponential relation with $T$ and number of agents.

To cope with the exponential relation associated with the number of timesteps, one can make the assumption that the environment can be treated as a finite Markov decision process (FMDP), in which the future states of the system depend only on the current state-action pair and not on the sequence of events before it. As will be seen in chapter 3, this simplification allows the possibility of creating a tensor network whose number of elements scale linearly with increasing $T$, rather than exponentially. 

In the case of the number of agents, linear algebra methods such as singular value decomposition can be utilised to find lower rank approximations of high order tensors \cite{SVD1}. When used appropriately this can keep calculations manageable, whilst maintaining a significant degree of accuracy. 

Another benefit of using TN methods is a set of optimisation techniques which can be used to optimise the policy with respect to the expected return, the fundamental concept of RL. These techniques are named as density matrix renormalisation group (DRMG) \cite{DMRG1,DMRG2,DMRG4}, and as with a large amount of the established TN methods, they were developed in the context of quantum physics but can be easily applied to TNs for other purposes.

A brief summary of the contents of this report: Chapter 2 contains an introduction to RL and FMDPs. Chapter 3 then formulates a TN to describe the expected return of a single agent FMDP. A practical application of this TN is shown in chapter 4, with an 1D random walker implementation.  Chapters 5 and 6 are analogous to 3 and 4, but for the case of multi-agent reinforcement learning. Finally, Chapter 7 concludes the report.
\chapter{Finite Markov Decision Processes}

\ifpdf
    \graphicspath{{Chapter2/Figs/Raster/}{Chapter2/Figs/PDF/}{Chapter2/Figs/}}
\else
    \graphicspath{{Chapter2/Figs/Vector/}{Chapter2/Figs/}}
\fi

In single-agent reinforcement learning problems there exists an agent that must achieve an objective within an environment. There are a number of variables that describe the dynamics of such a system. The first of these are the states, which describes the knowledge the agent has about the environment at a given timestep. States are described by the random variable, $S_{t}$, which can take values $s_{t}$ from a set $\mathcal{S}$. The second component are the actions, which describe the operations that the agent can perform in each timestep, to alter its state. The random variable in this case is $A_{t}$, taking values $a_{t}$ from a set $\mathcal{A}$. An agent that was initially in a state $s_{t-1}$ will take an action $a_{t-1}$ and move to a new state $s_{t}$. The agent will then obtain a reward, denoted by the random variable $R_{t}$. As before, this can take values $r_{t}$ from a set $\mathcal{R}$.  

At time $t=0$ the agent starts in a state $s_{0}$. The process described above is then repeated until some termination time $T$, in which no action is taken and the episode ends. One can describe the transitions using the following notation, which describes a trajectory between times $0$ and $T$:
\begin{equation}
    \omega_{0}^{T} = s_{0},a_{0} \xrightarrow[]{} s_{1},r_{1},a_{1} \xrightarrow[]{}...\xrightarrow[]{} s_{T-1},r_{T-1},a_{T-1}\xrightarrow[]{}s_{T},r_{T}
    \label{trajectory}
\end{equation}

The rewards are chosen in order to encode the objective that one wishes the agent to fulfill. The rewards for every timestep are summed into a quantity named the return, $G_{1:T}$. This is the assessment of the agents performance for the trajectory,

\begin{equation}
    G_{1:T} = \sum_{t=1}^{T} r_{t}
\end{equation}

One can also introduce the notion of the expected return, $\mathbb{E}(G_{1:T})$, which can be thought of as the average return that would be obtained if a very large quantity of trajectories were generated. 

The final concept to introduce is that of the policy. The policy, $\bm{\pi}$, is what determines what action the agent will take when in a given state at a given time. Importantly the actions taken in a given state can be different for each timestep. By indexing the policy at a given timestep, $\pi_{t}$ can also be written as a conditional probability of $A_{t-1}$ on $S_{t-1}$:

\begin{equation}
    \pi_{t}(a|s) = P(A_{t-1} = a_{t-1}|S_{t-1} = s_{t-1}) 
\end{equation}

As elements of the policy represent probabilities, they must be within the range $\pi_{t}(a|s)  \in  [0,1]$ and also be normalised so that the sum of probabilities adds up to one. This is the equivalent of stating that the agent must select an action in any state at each timestep,

\begin{equation}
    \sum_{a}\pi_{t}(a|s) = 1 \:\:  \forall\, s,t
\end{equation}

The task of reinforcement learning now becomes clear. Optimise $\bm{\pi}$ to maximise $\mathbb{E}(G_{1:T})$. Once the agent is fulfilling the desired objective as well as is possible (and therefore maximising $\mathbb{E}(G_{1:T})$), the policy is named the optimal policy, or $\bm{\pi}^{*}$. Equivalently:

\begin{equation}
    \bm{\pi}^{*} = \argmax_{\bm{\pi}}\mathbb{E}(G_{1:T}) 
\end{equation}

Referring back to equation \ref{trajectory}, $\omega_{0}^{T}$ represents the specific configuration of previous states, actions and rewards that was observed by the agent. Something that is important to consider is what the arrows, $\xrightarrow[]{}$, depict. Given that an agent has taken some trajectory $\omega_{0}^{t-1}$ and has just taken an action $a_{t-1}$, the following arrow represents how the agent transitions to state $s_{t}$ and obtains a reward $r_{t}$,
\begin{equation}
    \omega_{0}^{t-1}, a_{t-1} \xrightarrow[]{}s_{t},r_{t}
    \label{trajectory2}
\end{equation}

In the most general case, one must assume that $s_{t}$ and $r_{t}$ are dependant on $a_{t-1}$ as well as $\omega_{0}^{t-1}$. These transitions can then be written explicitly, as a conditional probability of $S_{t}$ and $R_{t}$ on the trajectory $\omega_{0}^{t-1}$ and the latest action $A_{t-1}$:
\begin{equation}
    P(S_{t} = s_{t},R_{t}= r_{t}|\Omega_{0}^{t-1} = \omega_{0}^{t-1},A_{t-1}=a_{t-1})
\end{equation}
The fact that this probability is conditional on the entirety of the trajectory causes simulations to be very computationally costly, as all previous states and actions must be recorded. For this reason, the assumption of Markov Decision Processes (MDPs) is often used. 

\section{Markov Decision Processes}
The defining factor of a MDP is the following. When in a state $S_{t-1}$ and taking an action $A_{t-1}$, the probability of transitioning to state $S_{t}$ and obtaining a reward $R_{t}$ depends only on the state and action taken at time $t-1$, and not on any previous states or actions in the trajectory. The transitions between states can be perfectly represented by the following conditional probability:

\begin{equation}
    P(S_{t} = s_{t}, R_{t} = r_{t}|S_{t-1} = s_{t-1}, A_{t-1} = a_{t-1}) = P_{t}(s',r|s,a)
    \label{markov}
\end{equation}

This probability doesn't depend on the full trajectory, meaning that simulations of MDPs are significantly less expensive. To extend this idea to a finite MDP (FMDP), a constraint is imposed  that the time, $t$, can only increase until a set value, $T$. On top of this, the sets $\mathcal{S}, \mathcal{A}$ and $\mathcal{R}$ are considered to have finite numbers of elements, which are $N_{S},N_{A}$ and $N_{R}$ respectively.


\section{Markov Matrices}
An important concept to introduce is that of Markov matrices, which will provide a useful platform to beginning the discussion of tensor networks. Consider a stripped back version of a FMDP, where actions and rewards are no longer considered, and the agent simply transitions between states depending on probabilities $P_{t}(s'|s)$.

At $t=0$ an agent starts in probability distribution of the possible values of $S_{0}$. This can be represented by a probability distribution vector, $\ket{{p}_{0}}$, which has a length equal to the total number of possible states, $N_{S}$. This vector can be decomposed into its basis coefficients and basis vectors by $\ket{{p}_{0}} = \sum_{s}c_{s}\ket{s}$. Each of  $\ket{s}$ satisfy the orthogonality relation of basis vectors, in a way that they are vectors of length $N_{S}$, with a one at index $s$ and zeros everywhere else. Importantly, the basis coefficients represent the probabilities of the agent being in a state, $c_{s} = P(S_{0}=s)$. The full vector has the form:

\begin{equation}
   \ket{{p}_{0}} = \begin{bmatrix}P(S_{0}=0),&..., & P(S_{0} = s-1), & P(S_{0} = s), &  ...&P(S_{0}=N_{S})\end{bmatrix}  
\end{equation}

Of course, as this vector represents a probability there exists that the condition that the sum over the elements must equal 1. If one wishes, the situation where the agent starts at a definitive state $S_{0} = \Tilde{s}$ can be depicted by making the value of the coefficients $c_{s} = \delta^{s,\tilde{s}}$.

To find $\ket{{p}_{1}}$, the probability distribution of what state the agent will be in at $t = 1$, the transition probabilities, $P(S_{1}=s_{1}|S_{0}=s_{0}) $, must be evaluated for all possible values of $S_{0}$ and $S_{1}$. If the system contains $N_{S}$ states, in total there will be ${N_{S}}^{2}$ of these probabilities. This information can be stored within a $N_{S}\times N_{S}$ Markov matrix, denoted by $\mathbf{M}_{1}$ \cite{markovmatrix}. The transition probability between two states, $s_{0}$ and $s_{1}$ is found in the $(s_{1},s_{0})$ element of $\mathbf{M}_{1}$. The form of such matrix is shown in equation \ref{M1}:

\begin{equation}
   \mathbf{M}_{1} =  \begin{bmatrix}
P(S_{1} = 0|S_{0} = 0)  & P(S_{1}=1|S_{0}=0)  & ... &P(S_{1}=N_{S}|S_{0}=0)\\
P(S_{1}=0|S_{0}=1) & P(S_{1}=1|S_{0}=1) & ...&P(S_{1}=N_{S}|S_{0}=1)
\\
\vdots & \vdots & \ddots&\vdots
\\P(S_{1}=0|S_{0}=N_{S})&P(S_{1}=1|S_{0}=N_{S})&...&P(S_{1}=N_{S}|S_{0}=N_{S})
\end{bmatrix}
\label{M1}
\end{equation}
Calculating $\ket{{p}_{1}}$ is then achieved by multiplying together the Markov matrix and $\ket{{p}_{0}}$:
\begin{equation}
    \ket{{p}_{1}} = \mathbf{M}_{1} \ket{{p}_{0}}
    \label{rank12tensors}
\end{equation}

To understand why this is the correct calculation, consider multiplying the first column of $\mathbf{M}_{1}$ by $\ket{{p}_{0}}$, which should yield the first element of $\ket{{p}_{1}}$ (equal to $P(S_{1}=0)$). Doing this multiplication yields:
\begin{equation}
\begin{split}
     & P(S_{1} = 0|S_{0} = 0)P(S_{0}=0) + P(S_{1} = 0|S_{0} = 1)P(S_{0}=1) +...\\
     & = \sum_{s}{P(S_{1}=0|S_{0} = s)}P(S_{0}=s) = P(S_{1} = 0)
\end{split}
\label{totalprobtheorem}
\end{equation}
Where the bottom line of equation \ref{totalprobtheorem} shows the use of the total probability theorem \cite{totalprobtheorem}. 

 In the calculation shown in equation \ref{rank12tensors}, one can write the kets in vector form, so that the index can be shown explicitly. It is then very clear that a contraction was used, with the shared index being $s_{0}$. This can be displayed graphically, as seen below in equation \ref{p0m1}:

\begin{equation}
    \Vec{p}_{1}^{\,\,s_{1}} = \sum_{s_{0}} \mathbf{M}_{1}^{s_{1},s_{0}}\Vec{p}_{0}^{\,\,s_{0}}
\end{equation}

\begin{equation}
    \includegraphics[width = 6cm]{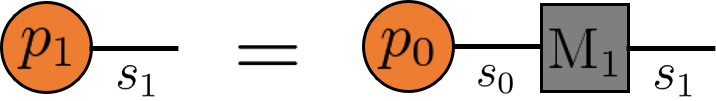}
    \label{p0m1}
\end{equation} It can be seen that the $\mathbf{M}_{1}$ tensor has a rank of 2, but is displayed as a square. This is because this tensor will be built up through the next section, until it is rank 4.

The same process can be used to obtain $\ket{{p}_{2}}$, except one now uses a different Markov matrix, $\mathbf{M}_{2}$, as it cannot be assumed that the transition probabilities are equivalent for every time step. First writing the formula for $\ket{{p}_{2}}$, one can make a substitution for $\ket{{p}_{1}}$ using equation \ref{rank12tensors}:

\begin{equation}
    \ket{{p}_{2}} = \mathbf{M}_{2} \ket{{p}_{1}}=\mathbf{M}_{2} \mathbf{M}_{1} \ket{{p}_{0}}
\end{equation}

A clear pattern has now emerged. To find the probability state vector at a timestep $T$, in terms of the initial probability state vector, $\ket{{p}_{0}}$, the following formula is used:

\begin{equation}
    \ket{{p}_{T}} = \mathbf{M}_{T}\mathbf{M}_{T-1}... \mathbf{M}_{2} \mathbf{M}_{1}\ket{{p}_{0}}
    \label{mps}
\end{equation}

\begin{equation}
    \includegraphics[width = 12cm]{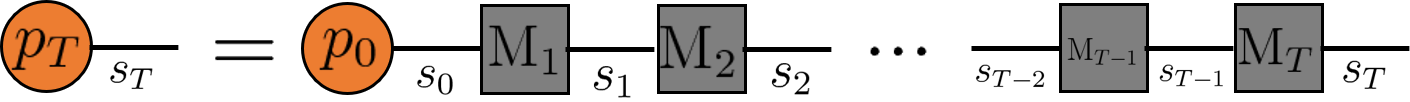}
    \label{pt}
\end{equation}
The above structure is termed a Markov chain \cite{markovchain}.This fundamental result will prove very useful in the following chapter.

\chapter{Tensor Networks for FMDPs} \label{chap3}

\ifpdf
    \graphicspath{{Chapter3/Figs/Raster/}{Chapter3/Figs/PDF/}{Chapter3/Figs/}}
\else
    \graphicspath{{Chapter3/Figs/Vector/}{Chapter3/Figs/}}
\fi

The previous chapter concluded with an explanation of how to propagate the probability state vector through time. This chapter will focus on extending these ideas to finding a tensor network that is capable of representing the expected return of a FMDP. Before considering actions and rewards, there is one more concept that must be introduced, which is the flat tensor.

 One may contract the remaining index in equation \ref{pt}, $s_{T}$, with the flat tensor, $\ket{-_{s}}$. This tensor is a vector of length $N_{S}$ that has ones at every element. When contracted with an index, this has the effect of marginalising the elements of the vector. In this case, the result of contracting the tensor network is the integer 1, as the elements of the probability state vector $\ket{{p}_{T}}$ have been summed over. This is shown in equations \ref{mpsflat} and \ref{flattensor}, where the flat tensor is represented by a line perpendicular to the index protruding from $\mathbf{M}_{T}$:

    \begin{equation}
    \braket{-_{s}|{p}_{T}} = \bra{-_{s}}\mathbf{M}_{T}\mathbf{M}_{T-1}... \mathbf{M}_{2} \mathbf{M}_{1}\ket{{p}_{0}} = 1
    \label{mpsflat}
\end{equation}

\begin{equation}
    \includegraphics[width = 11cm]{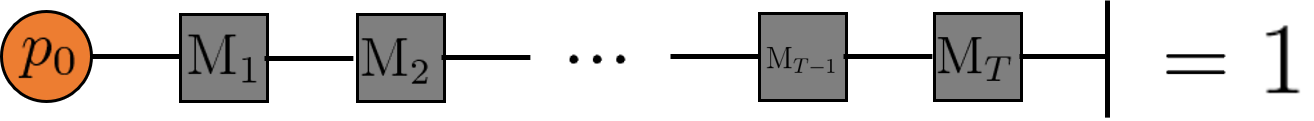} 
    \label{flattensor}
\end{equation}

Until now, the rank 2 $\mathbf{M}_{t}$ tensor has represented the relation between the two variables $S_{t}$ and $S_{t-1}$, and more specifically the transitions: $P_{t}(s'|s)$. In the setting of a FMDP, actions and rewards also must be considered. This means that there are four separate components in vector space: the previous state $S_{t-1}$, the next state $S_{t}$, the actions $A_{t-1}$ and the rewards $R_{t}$. To find the transition probabilities between these components, the rank of the $\mathbf{M}_{t}$ tensor (which is currently 2) must be increased to 4. This proves to be fairly involved, so in this chapter the $\mathbf{M}_{t}$ tensor will be built up progressively. The rewards will be introduced first, so that the rank of the $\mathbf{M}_{t}$ tensor is increased to 3, and will represent the transitions $P_{t}(s',r|s)$. The rewards are added first, due to the fact that these probabilities have the form of those associated with the Hidden Markov Model (HMM). Finally, the actions will be added, to create the full rank 4 tensor that perfectly describes transitions in FMDPs, $P_{t}(s',r|s,a)$. 

\section{Tensor Network with Rewards}
In accordance with the discussion on FDMPs, the agent must be given a reward in each timestep. To do this, an extra component is being added in vector space. This means the $\mathbf{M}_{t}$ tensor now has to find a relation between three variables, and therefore the rank must increase by one, so that it now has the form:  
\begin{equation}
    \includegraphics[width=5cm]{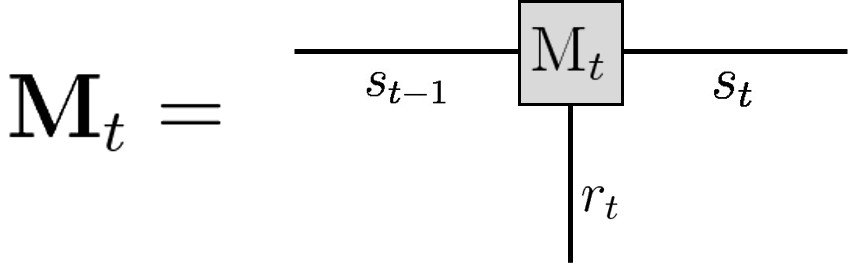}
    \label{Mwithrewards}
\end{equation}

Note that the change in colour of the tensor represents that it is now representing a different conditional probability. The conditional probability represented by the tensor in equation \ref{Mwithrewards} is $P(s_{t},r_{t}|s_{t-1})$, and it can be written with its indices as $\mathbf{M}_{t}^{s_{t},r_{t},s_{t-1}}$. It is related to the rank 2 $\mathbf{M}_{t}$ tensor via marginalisation:
\begin{equation}
    \sum_{r_{t}} \mathbf{M}_{t}^{s_{t},r_{t},s_{t-1}} = \mathbf{M}_{t}^{s_{t},s_{t-1}}
\end{equation}
When $\mathbf{M}_{t}$ was rank 2, the state transitions were completely unknown to the agent. However, now the agent receives pieces of information about the system, in the form of rewards. This is in direct analogy to the Hidden Markov Model (HMM) \cite{hmm}. If in equation \ref{Mwithrewards}, $s_{t-1}$ and $s_{t}$ were contracted, the remaining $r_{t}$ index would represent a vector of the probabilities of observing each reward. It can be written as $\ket{p_{r,t}}$, and can be decomposed in terms of basis vectors in the same way as the states: $\ket{p_{r,t}} = \sum_{r}c_{r,t}\ket{r}$. As before, $c_{r,t} = P(R_{t}=r)$.

Consider now a tensor network like the one seen in equation \ref{flattensor}, but formed with the new rank 3 $\mathbf{M}_{t}$ tensors for the case $T=4$,

\begin{equation}
 \includegraphics[width = 9cm]{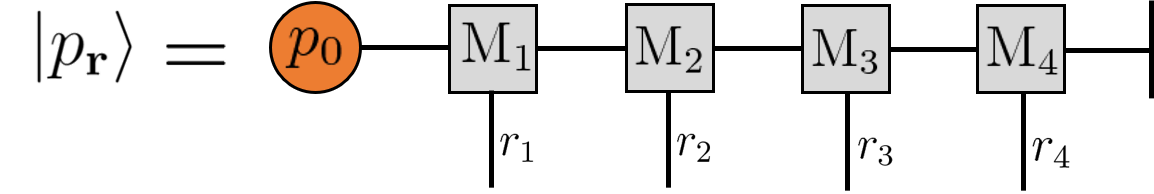}
    \label{fourreward}
\end{equation}
This tensor network has four indices that are not contracted, and the resulting probability distribution, $\ket{p_{\mathbf{r}}}$ is calculated by taking the tensor product of $\ket{p_{r,t}}$ for all $t$,

\begin{equation}
    \ket{p_{\mathbf{r}}} = \bigotimes_{t=1}^{T}\ket{p_{r,t}}
\end{equation}
In the case of $T=4$ (as in equation \ref{fourreward}), $\ket{p_{\mathbf{r}}}$ represents a probability $P(r_{1},r_{2},r_{3},r_{4})$. The objective is now to find a tensor that when contracted with the indices $r_{1},r_{2},r_{3}$ and $r_{4}$, gives the expected return, $\mathbb{E}(G)$. This tensor can be named $\mathbf{G}$.
In this case a matrix product operator can be used to determine the form of the tensor. 

\subsubsection{Matrix Product Operator}
One must first introduce the reward operator, $\hat{R}$. When this acts on one of the reward basis vectors $\ket{r}$, it returns an eigenvalue that is equal to the value of the reward,

\begin{equation}
    \hat{R}\ket{r} = r\ket{r}
\end{equation}

The operator $\hat{R}$ should be thought of in matrix form, where the bra and ket simply correspond to indices of the matrix. Denoting the possible values of $r$ as $r^{(0)}, r^{(1)}...$, the matrix is shown below:

\begin{equation}
    \hat{R} = \begin{bmatrix}
\bra{r=r^{(0)}}\hat{R}\ket{r=r^{(0)}} & \bra{r=r^{(1)}}\hat{R}\ket{r=r^{(0)}} &...\\
\bra{r=r^{(0)}}\hat{R}\ket{r=r^{(1)}} & \bra{r=r^{(1)}}\hat{R}\ket{r=r^{(1)}} & ...\\
\vdots & \vdots & \ddots
\end{bmatrix}
\label{Rmatrix}
\end{equation}

As previously mentioned, $\ket{r}$ are orthogonal basis vectors. Therefore they obey the relation $\braket{r^{(i)}|r^{(j)}} = \delta^{i,j}$. Using this allows for the simplification of the $\hat{R}$ matrix:

\begin{equation}
    \hat{R} = \begin{bmatrix}
r^{(0)} & 0 &...\\
0& r^{(1)}& ...\\
\vdots & \vdots & \ddots
\end{bmatrix}
\end{equation}

When this reward operator acts on the probability reward vector, $\ket{p_{r,t}}$, it essentially acts to multiply each of the coefficients, $c_{r,t}$, by the value of the reward, $r$. In this case, the eigenvalue is a vector of the possible values of rewards, denoted as $\vec{r}$. This is simple to show:

\begin{equation}
    \hat{R} \ket{p_{r,t}} = \sum_{r}c_{r,t}\hat{R}\ket{r} = \sum_{r}c_{r,t}r\ket{r} = \vec{r} \cdot \ket{p_{r,t}}
\end{equation}
By also applying the flat reward tensor, $\ket{-_{r}}$, (equivalent to the flat state tensor but with $N_{R}$ elements), one obtains the expected reward for the timestep $t$:

\begin{equation}
   \bra{-_{r}} \hat{R} \ket{p_{r,t}} =\bra{-_{r}} \vec{r} \cdot \ket{p_{r,t}} = \sum_{i}r^{(i)}P(r=r^{(i)}) = \mathbb{E}(R_{t})
\end{equation}

One also defines a return operator as the following: 

\begin{equation}
    \hat{G}_{1:T} = \sum_{t=1}^{T} \hat{R}_{t}     
    \label{Gnontensor}
\end{equation}

where $\hat{R}_{t}$ is exactly equal to the previously defined $\hat{R}$ operator, and it acts on the rewards at timestep $t$. $\hat{G}_{1:T}$ can also be described using a tensor as $\hat{R}$ was in equation \ref{Rmatrix}. For the case shown in figure \ref{fourreward}, the form of the tensor is found by tensor product of the reward operators \cite{DMRG2}, which yields a rank 8 tensor:

\begin{equation}
\begin{split}
    \hat{G}_{1:4} & =  \hat{R}_{1}\otimes  \hat{\mathbb{I}} \otimes  \hat{\mathbb{I}} \otimes  \hat{\mathbb{I}} \\
    & +  \hat{\mathbb{I}} \otimes \hat{R}_{2} \otimes  \hat{\mathbb{I}} \otimes  \hat{\mathbb{I}} \\
    & +  \hat{\mathbb{I}}  \otimes  \hat{\mathbb{I}} \otimes \hat{R}_{3}\otimes  \hat{\mathbb{I}} \\
    & +  \hat{\mathbb{I}}  \otimes  \hat{\mathbb{I}} \otimes  \hat{\mathbb{I}} \otimes \hat{R}_{4}\\
\end{split}
    \label{Gnontensor}
\end{equation}
where $\hat{\mathbb{I}}$ is the identity operator, in the form of a matrix of the same dimensions as $\hat{R}_{t}$. When acting $\hat{G}_{1:4}$ on $\ket{p_{\mathbf{r}}}$, it is clear that $\hat{R}_{1}$ will act on $\ket{p_{r,1}}$, $\hat{R}_{2}$ will act on $\ket{p_{r,2}}$ and so on.

One problem with the form of the $\hat{G}_{1:4}$ tensor, is that its rank scales linearly with increasing $T$. To avoid this, one can decompose the tensor into smaller ones, so that there is one tensor per timestep. To do this, new tensors $\mathbf{w}_{1}, \mathbf{W}_{2}, \mathbf{W}_{3}, $ and $\mathbf{w}_{4} $ are introduced, so that the tensor $\mathbf{G}$ can be written as:
\begin{equation}
    \mathbf{G}_{1:4} =\mathbf{w}_{1}\mathbf{W}_{2}\mathbf{W}_{3}\mathbf{w}_{4} 
    \label{workoutoperators}
\end{equation}

\begin{equation}
    \includegraphics[width=13cm]{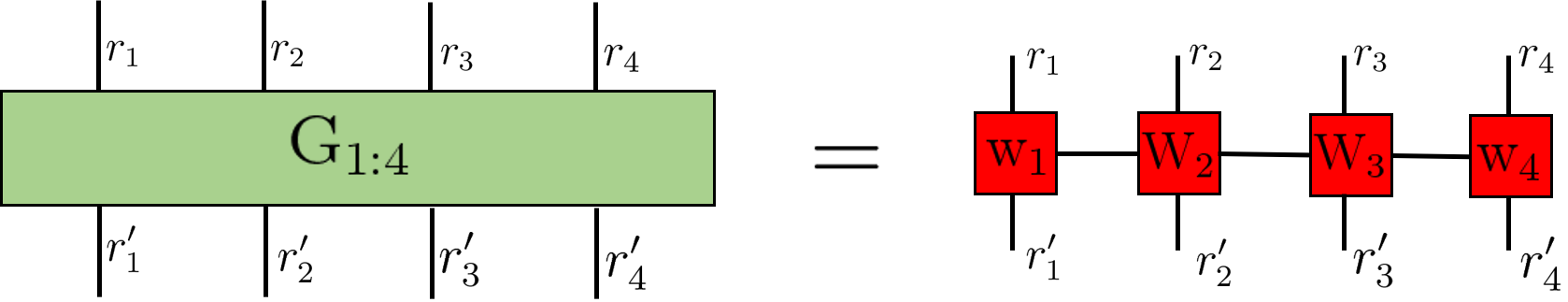}
\end{equation}

Note that in the general case, both $\mathbf{w}_{1}$ and $\mathbf{w}_{T}$ are rank 3 tensors but should be thought of as $2\times 1$ vectors, with operators as the elements. Similarly the $\mathbf{W}_{t}$ tensor is a rank 4 tensor, but should be thought of as a $2\times2$ matrix (again with operators as the elements). Each operator is represented by a $N_{R}\times N_{R}$ matrix. Without knowing the actual operators, $\mathbf{w}_{1}$, $\mathbf{w}_{4}$ and $\mathbf{W}_{t}$ can be written as:
\begin{equation}
    \mathbf{w}_{1} = \begin{bmatrix}
\hat{\mathbb{A}} & \hat{\mathbb{B}}

\end{bmatrix}
\end{equation}
\begin{equation}
    \mathbf{w}_{4} = \begin{bmatrix}
\hat{\mathbb{C}}\\
\hat{\mathbb{D}}
\end{bmatrix}
\end{equation}

\begin{equation}
    \mathbf{W}_{t} = \begin{bmatrix}
\hat{\mathbb{F}} & \hat{\mathbb{J}}\\
\hat{\mathbb{H}} & \hat{\mathbb{L}}
\end{bmatrix}
\end{equation}
When multiplying these objects together, the elements are multiplied by tensor product. For example:

\begin{equation}
    \mathbf{w}_{1}\mathbf{W}_{2} = \begin{bmatrix}
    \hat{\mathbb{A}}\otimes \hat{\mathbb{F}} + \hat{\mathbb{B}}\otimes \hat{\mathbb{H}} & \hat{\mathbb{A}}\otimes \hat{\mathbb{J}} +
    \hat{\mathbb{B}}\otimes \hat{\mathbb{L}}
    \end{bmatrix}
\end{equation}

One can now carry out the calculation shown in equation \ref{workoutoperators}, and then can compare the result to \ref{Gnontensor} to attempt to work out the operators in the $\mathbf{w}_{1}$, $\mathbf{w}_{4}$ and $\mathbf{W}_{t}$ tensors:

\begin{equation}
\begin{split}
    \mathbf{G}_{1:4} & = \left((\hat{\mathbb{A}}\otimes \hat{\mathbb{F}} + \hat{\mathbb{B}}\otimes\hat{\mathbb{H}})\otimes\hat{\mathbb{F}} + (\hat{\mathbb{A}}\otimes\hat{\mathbb{J}}+\hat{\mathbb{B}}\otimes\hat{\mathbb{L}})\otimes\hat{\mathbb{H}}\right)\otimes\hat{\mathbb{C}} \\
    & + \left((\hat{\mathbb{A}}\otimes \hat{\mathbb{F}} + \hat{\mathbb{B}}\otimes\hat{\mathbb{H}})\otimes\hat{\mathbb{J}} + (\hat{\mathbb{A}}\otimes\hat{\mathbb{J}}+\hat{\mathbb{B}}\otimes\hat{\mathbb{L}})\otimes\hat{\mathbb{L}}\right)\otimes\hat{\mathbb{D}}
\end{split}
\end{equation}

For reasons that will become clear,  $\mathbb{J}$ is set to a 'zero' operator (a matrix of zeros). The equation simplifies to the following:
\begin{equation}
\begin{split}
    \mathbf{G}_{1:4} = \hat{\mathbb{A}}\otimes \hat{\mathbb{F}}\otimes\hat{\mathbb{F}}\otimes\hat{\mathbb{C}} + \hat{\mathbb{B}}\otimes \hat{\mathbb{H}}\otimes\hat{\mathbb{F}}\otimes\hat{\mathbb{C}} + \hat{\mathbb{B}}\otimes \hat{\mathbb{J}}\otimes\hat{\mathbb{H}}\otimes\hat{\mathbb{C}} +\hat{\mathbb{B}}\otimes \hat{\mathbb{L}}\otimes\hat{\mathbb{L}}\otimes\hat{\mathbb{D}}
\end{split}
\end{equation}

Comparing this to \ref{Gnontensor}, it is clear that $\hat{\mathbb{A}}, \hat{\mathbb{H}}, \hat{\mathbb{D}} = \hat{R}$ and $\hat{\mathbb{B}}, \hat{\mathbb{C}}, \hat{\mathbb{F}}, \hat{\mathbb{L}} = \hat{\mathbb{I}}$. To summarise, the final forms of $\mathbf{w}_{1}$, $\mathbf{w}_{4}$ and $\mathbf{W}_{t}$ are as follows:
\begin{equation}
    \mathbf{w}_{1} = \begin{bmatrix}
\hat{R} & \hat{\mathbb{I}}

\end{bmatrix}
\end{equation}
\begin{equation}
    \mathbf{w}_{4} = \begin{bmatrix}
\hat{\mathbb{I}}\\
\hat{R}
\end{bmatrix}
\end{equation}

\begin{equation}
    \mathbf{W}_{t} = \begin{bmatrix}
\hat{\mathbb{I}} & \hat{0}\\
\hat{R} & \hat{\mathbb{I}}
\end{bmatrix}
\end{equation}

Equation \ref{expectedreturnnoaction} shows how these tensors are integrated into the tensor network, using the flat reward tensors to sum over the other index, $r'$. The result is the expected return:
\begin{equation}
    \includegraphics[width=10cm]{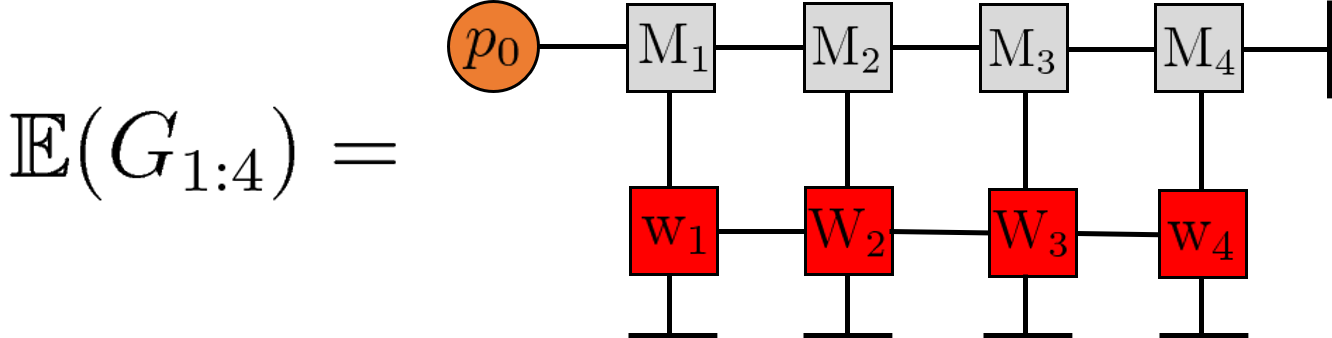}
    \label{expectedreturnnoaction}
\end{equation}

To make one more slight simplification, the flat reward tensors can be contracted with the $\mathbf{W}$ tensors to create new tensors $\Tilde{\mathbf{W}}$. Importantly, this makes the elements of the tensors vectors rather than matrices (that represented operators). The element corresponding to $\hat{R}$ now contains a vector with elements corresponding to the values of the rewards, $\vec{R}$. The element corresponding to $\hat{\mathbb{I}}$ is a vector with all elements being a one, $\vec{1}$ and the element corresponding to the zero operator is now a vector filled with zeros, $\vec{0}$.
The forms of the new tensors are the following:

\begin{equation}
    \Tilde{\mathbf{w}}_{1} = \begin{bmatrix}
\vec{R} & \vec{1}

\end{bmatrix}
\end{equation}
\begin{equation}
    \Tilde{\mathbf{w}}_{4} = \begin{bmatrix}
\vec{1}\\
\vec{R}
\end{bmatrix}
\end{equation}

\begin{equation}
    \Tilde{\mathbf{W}}_{t} = \begin{bmatrix}
\vec{1} & \vec{0}\\
\vec{R} & \vec{1}
\end{bmatrix}
\end{equation}

Finally, a TN using these new tensors is shown in equation \ref{tilde}.

\begin{equation}
    \includegraphics[width=10cm]{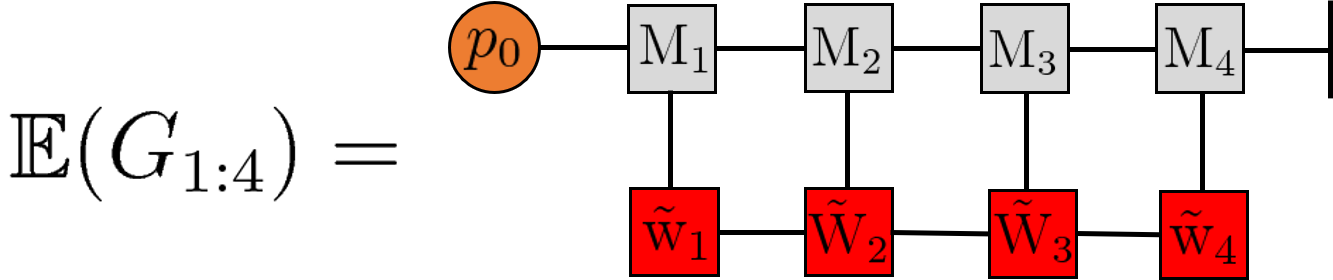}
    \label{tilde}
\end{equation}

Importantly, the $\Tilde{\mathbf{W}}_{t}$ tensor's rank does not increase with $T$ but instead, a new tensor is introduced each time. This structure avoids the curse of dimensionality.
\section{Tensor Network with Actions}
Now that the addition of the rewards is complete, the agent receives an assessment of its performance, without having any control over its state. The actions are the final component that must be considered to formulate the complete TN representation of FDMPs. One must increase the rank of the $\mathbf{M}_{t}$ tensors to 4, as they now represent the relation between four components, $\mathbf{M}_{t}^{s_{t},r_{t},s_{t-1},a_{t-1}} = P(s_{t},r_{t}|s_{t-1},a_{t-1})$.

To add the actions to the TN representation, the policy must also be considered, as it is the policy that generates the actions based on the previous state:
\begin{equation}
    \pi_{t}(a|s) = P(A_{t-1} = a_{t-1}|S_{t-1} = s_{t-1})
\end{equation}

It can be seen from this conditional probability that the policy relates two variables in vector space: the states and the actions. Following from the previous discussion in section 3.1, this means a rank 2 tensor will be capable of representing it, which can be expressed as $\bm{\pi}^{a_{t-1},s_{t-1}}_{t}$. When contracting the policy with the rank 4 $\mathbf{M}_{t}$ tensor, it may appear that there are two shared indices, both the previous state $s_{t-1}$ and the previous action $a_{t-1}$. In this case one may expect to contract both of the indices, and to be left with a rank 2 tensor, with indices for the next state $s_{t}$ and the next reward $r_{t}$. However this is not correct. The reason for this is that the $s_{t-1}$ index represents a conditional variable in both $\bm{\pi}_{t}$ and $\mathbf{M}_{t}$ (it is on the right hand side of the conditional symbol in both $P(s_{t},r_{t}|s_{t-1},a_{t-1})$ and $P(a_{t-1}|s_{t-1})$), and therefore cannot be contracted \cite{conditional,tensornotation2}. The correct resulting object when contracting these tensors is a rank 4 tensor, with its formation shown in equation \ref{B} and graphically in equation \ref{Btensor}:

\begin{equation}
   \sum_{a_{t-1}} \mathbf{M}^{s_{t},r_{t},s_{t-1},a_{t-1}}_{t}\bm{\pi}^{a_{t-1},s_{t-1}}_{t} = \mathbf{B}^{\,s_{t},r_{t},s_{t-1},s_{t-1}}_{t}
   \label{B}
\end{equation}

\begin{equation}
     \includegraphics[width = 9cm]{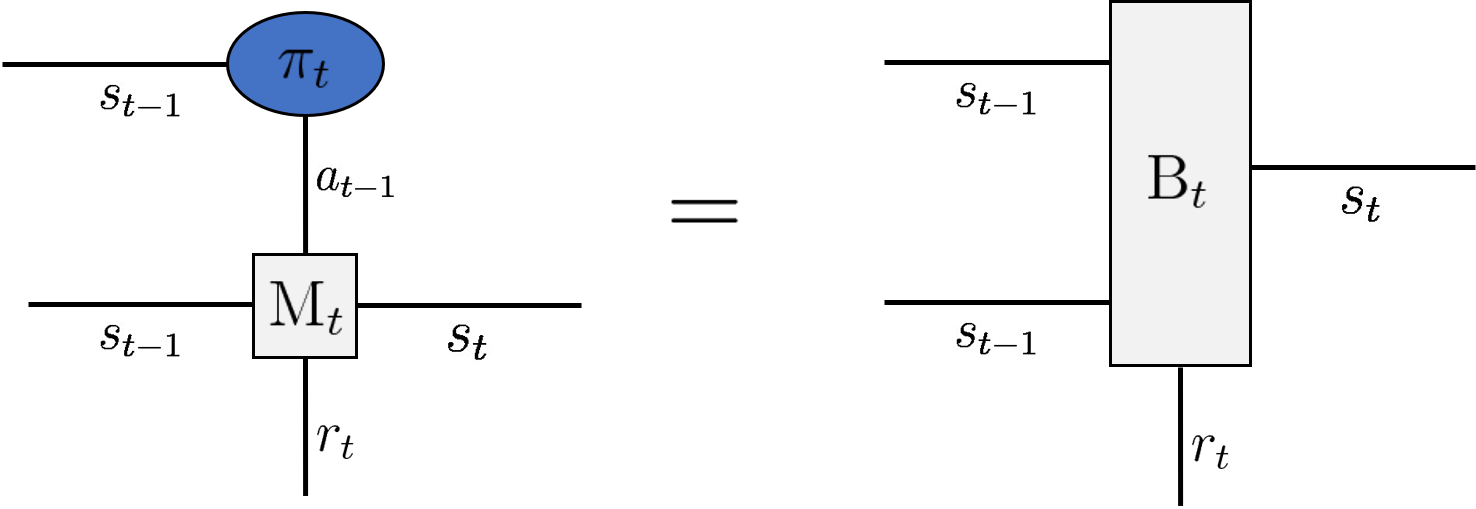}\label{Btensor}
\end{equation}
   
If this process of contracting $\bm{\pi}_{t}$ and $\mathbf{M}_{t}$ was repeated for every timestep, then there would exist a $\mathbf{B}_{t}$ tensor for every $t$, except for at $t=0$ (where the initial probability state vector, $\ket{p_{0}}$, exists). With current knowledge however, it would not be possible to contract the entire TN. This is because in each $\mathbf{B}_{t}$, there are two $s_{t-1}$ indices and only one $s_{t}$ index coming out which means the states cannot be contracted together.  

\subsubsection{Copy Tensor}
To solve this problem, a copy tensor is introduced. When presented with a vector of an input, $s$, this type of tensor has the ability to produce copies of this input at its other indices \cite{copy}. To display this as a conditional probability, a rank 3 copy tensor can be represented by the following equation:

\begin{equation}
    P(s',s''|s) =
    \begin{cases}
    1, & \text{if}\ s = s' = s'' \\
    0, & \text{otherwise}\\
    \end{cases}
\end{equation}

By writing it as a tensor with its indices, one can understand that a copy tensor is in fact just a multidimensional identity array, with ones on the main diagonal and zeros everywhere else,

\begin{equation}
    \bm{\Delta}^{s,s',s''} =
    \begin{cases}
    1, & \text{if}\ s= s' = s'' \\
    0, & \text{otherwise}\\
    \end{cases}
\end{equation}

Finally, in graphical equations the copy tensor is represented as a dot:

\begin{equation}
 \includegraphics[width = 6cm]{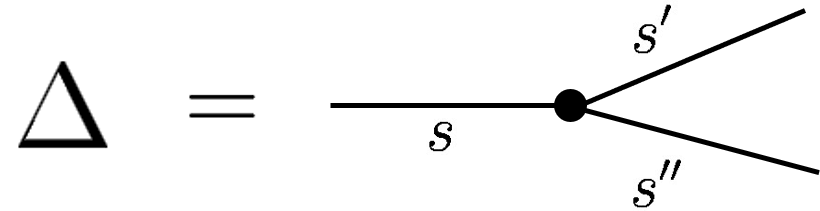}
\end{equation}

The copy tensor enables the capability of contracting together the $\mathbf{B}_{t}$ tensors, and also contracting the $\ket{{p}_{0}}$ vector with $\mathbf{B}_{1}$. An example showing how $\mathbf{B}_{t}$ and $\mathbf{B}_{t+1}$ are contracted is shown in equation \ref{contractb}, creating a new object, $\mathbf{B}_{t:t+1}$, that describes state transitions between $t$ and $t+1$. This can then be contracted with the tensors for larger $t$ until all tensors have been contracted ($t = T$),

\begin{equation}
    \includegraphics[width=12cm]{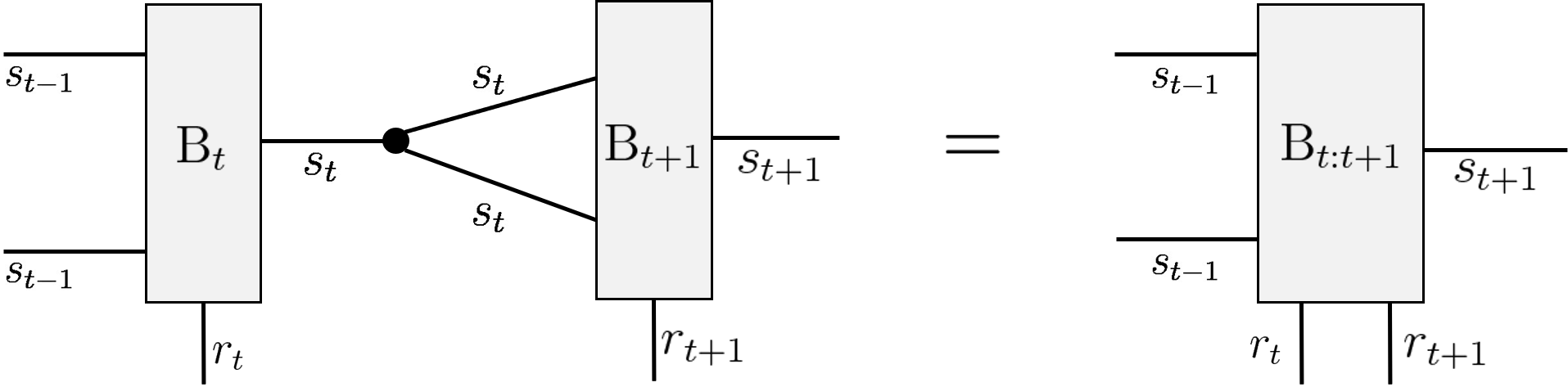}
    \label{contractb}
\end{equation}

The result of contracting the entire tensor network is once again the expected return, $\mathbb{E}(G_{1:T})$. By converting the $\mathbf{B}_{t}$ tensors back to the expression of $\mathbf{M}_{t}$ and $\bm{\pi}_{t}$, the full tensor network for a FMDP is shown for $T=4$ in equation \ref{finalactiontensor}:

\begin{equation}
    \includegraphics[width = 13cm]{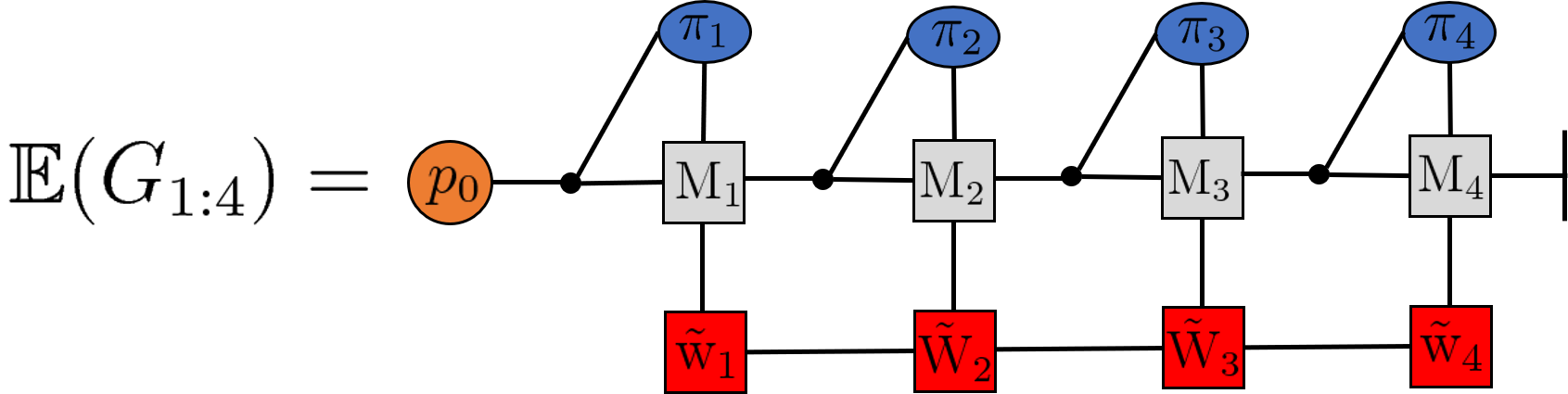}
    \label{finalactiontensor}
\end{equation}

\section{Policy Optimisation}

 The fundamental concept of RL is optimising the policy to maximise $\mathbb{E}(G_{1:T})$. To do this within the TN setting, one can use existing optimisation techniques that have already been designed for other applications of TNs. One algorithm that is used regularly is the density matrix re-normalisation group algorithm (DMRG). The core of the algorithm, is to optimise the variational tensors within the network with respect to an objective function. Optimising all the variational tensors at once can prove to be expensive, as the cost will grow exponentially with number of tensors being optimised. To avoid this problem, the variational tensors can be optimised one at a time, under the assumption that the rest of the tensor network is fixed \cite{DMRG2,singlesitedmrg}. 

In the tensor network that has been developed for FMDPs in this chapter, the objective function is the expected return, $\mathbb{E}(G_{1:T})$, and the variational parameters that one wishes to optimise are the policy tensors for each timestep. To understand how to practically optimise one of these tensors, consider the $\bm{\pi}_{3}$ tensor in equation \ref{finalactiontensor}. One must first contract the entirety of the rest of the network. What remains is a rank 2 tensor that is being contracted with $\bm{\pi}_{3}$, and this is seen in figure \ref{dmrgcontract}:

\begin{figure}[h]
    \centering
    \includegraphics[width=15cm]{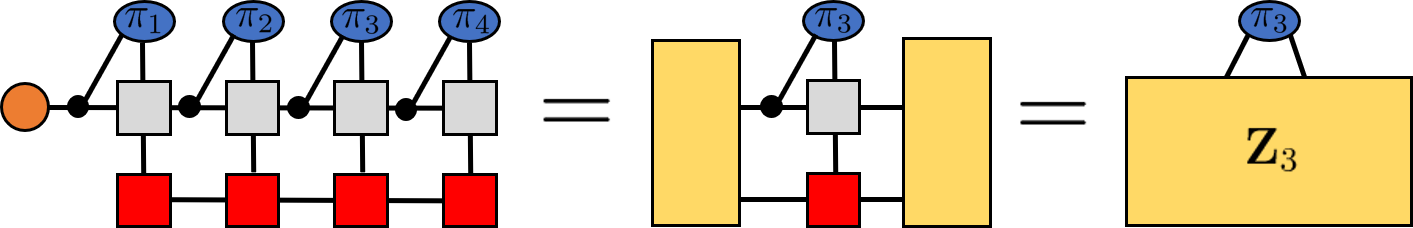}
    \caption{Showing how the tensor network is contracted leaving $\bm{\pi}_{3}$ being contracted with another object termed $\mathbf{Z}_{3}$. Note the order of contraction, which involves contracting the first two layers, and the last layer individually. These layers are then contracted with the third layer. This order is superior to just contracting in time order, as it ensures that the ranks of resulting tensors are minimised at any time throughout the contractions.}
    \label{dmrgcontract}
\end{figure}

This tensor can be labelled $\mathbf{Z}_{3}$, and has indices for $S_{t-1}$ and $A_{t-1}$ (as does the policy). The value at the $s,a$ index corresponds to the expected return given that the agent takes an action $a$ in state $s$. The objective is to now alter the values in $\bm{\pi}_{3}$ so that when contracted with $\mathbf{Z}_{3}$ the resulting scalar is maximized (as this scalar represents $\mathbb{E}(G_{1:4})$). Of course when doing this, one must remember that the values in $\bm{\pi}_{t}$ represent probabilities and therefore must obey the constraints $\sum_{a}\bm{\pi}_{t}^{a,s} = 1 \, \forall \, s,t$ and $\bm{\pi}_{t}^{a,s} \in [0,1]$. The simplest way to optimise the policy is to simply make it greedy with respect to the other tensor. In practice this means looping through all possible values of $s$ in $\mathbf{Z}_{3}$ and for each, finding which action causes the largest expected return. Then one updates the policy by making the probability of this action (with the largest expected return) equal to 1, and the probability of all other actions 0. If all actions give the same expected return, then one does not need to change to values in the policy. This can be written mathematically by first finding the greedy action at a particular state $s$:
\begin{equation}
    a' = \argmax_{a} [\bm{\pi}_{3}^{s} \cdot \mathbf{Z}_{3}^{s}]
\end{equation}
Then updating the policy in the following way:

\begin{equation}
    \bm{\pi}_{3}^{a,s} = \delta^{a,a'}
\end{equation}

It has also been shown that in this single agent case, if one begins by optimising the policy at the last timestep, and then working backwards through time, this will always converge to the optimal policy \cite{edpaper}. In the example shown in figure \ref{finalactiontensor}, one would optimise the policies in the following order $\bm{\pi}_{4}$, $\bm{\pi}_{3}$, $\bm{\pi}_{2}$, $\bm{\pi}_{1}$. 

\section{Final Remarks}
It should now be clear why the approximation of a FMDP is essential for the use of TNs within RL. At a given timestep $t = t'$, if the state and reward depended on the entirety of the states accessed by the agent before this time, the $\mathbf{M}_{t'}$ tensor would have to have an index for each value of the state at every previous timestep. This would of course mean that the rank of the $\mathbf{M}_{t}$ tensors would increase linearly with $t$. A linear increase in rank causes an exponential increase in the number of parameters that are stored, and operations that are performed when the tensors are contracted. Consequently, it will not take many timesteps before the computation becomes too large for any computer to be able to process it. For example, if $N_{S} = 20, T = 20, N_{A} = 2$ and $N_{R} = 3$, the number elements in the $\mathbf{M}_{t = 20}$ tensor would be equal to $20^{20}\times 2 \times 3 \approx 10^{26}$, which would take far more memory to store than is available in any existing computer \cite{supercomputer}. Equation \ref{finalactiontensor} shows a structure, where the computational time required to contract it scales linearly with number of timesteps. This indicates how the assumption of a FMDP helps avoid the curse of dimensionality.



\chapter{SARL Example: 1D Random Walker}

\ifpdf
    \graphicspath{{Chapter4/Figs/Raster/}{Chapter4/Figs/PDF/}{Chapter4/Figs/}}
\else
    \graphicspath{{Chapter4/Figs/Vector/}{Chapter4/Figs/}}
\fi

The 1D random walker is a much studied problem due to its simplicity, and its applications all over physics, ranging from nano-science to quantum field theory \cite{nanoscience,qft}. On top of this, it is often the example used to demonstrate simple RL in the context of FMDPs \cite{randomwalk,randomwalk2}.

In this specific example, there is an agent which starts at $S_{t=0} = 0$. In each timestep the agent must either take an action to go up or down one step, so that $A_{t} \in \{1,-1\}$. The agent continues until $t$ reaches the termination time $T$. One can see that the total possible number of states is $N_{S} = 2T+1$, where the the state with the largest number ($S = T$) can only be accessed when taking the action to go up in every timestep. In the purely deterministic setting, the transitions between states can be described using the following formula:

\begin{equation}
    S_{t+1} = S_{t} + A_{t}
    \label{nonoise}
\end{equation}
The 1D random walker starts with a policy that reflects its randomness. This involves an equal probability for the agent to go up or down in each timestep:

\begin{equation}
    \pi_{t}(a|s) = 0.5 \:\: \forall\:\: a, s, t 
    \label{randompolicy}
\end{equation}

\section{Rewards}
By choosing the rewards, an objective is encoded within the simulation. For this example it was decided that the objective would be that the agent would stay above $S = -1$ for all timesteps, and at $t=T$, the agent would return to $S = 0$. This is written explicitly in equation \ref{objective}:

\begin{equation}
    \textrm{OBJECTIVE} = \:\:\: 
    \begin{cases}
    S_{t}\geq 0, & \text{if}\ 1\leq t <T \\
    S_{t} = 0, & \text{if}\ t = T
    \end{cases}
    \label{objective}
\end{equation}

To encode this objective, the following rewards were formulated: \begin{align*}
    \textrm{IF}\:\: t<T: \\
    r_{t} =
    \begin{cases}
    0, & \text{if}\ S_{t} \geq 0\\
    -1, & \text{otherwise}
    \end{cases}\\
    \\
    \textrm{ELSE IF}\:\: t=T: \\
    r_{t} =
    \begin{cases}
    1, & \text{if}\ S_{t} = 0\\
    -10, & \text{otherwise}
    \end{cases}\\
\end{align*}
Of course there are many other suitable values for the rewards to encode the same objective. The only thing that matters is the relative differences between rewards for given states. One final thing to clarify, is that with this environment, the $\mathbf{M}_{t}$ tensors are identical for all $t<T$. 



\section{Random Noise}
To make the simulation less deterministic, random noise was added. This causes an uncertainty in the resulting state, $S_{t+1}$, when taking an action $A_{t}$ in a state $S_{t}$. To do this, a discrete random variable, $\mathcal{X}_{d}$, is added to the action to create a new action, $\Tilde{A}_{t}$. It was chosen that this random variable would belong to a  discretised normal distribution. The resulting change to the state in each timestep is given by this modified version of equation \ref{nonoise}:

\begin{equation}
    S_{t+1} = S_{t} + \Tilde{A}_{t} = S_{t} + A_{t} + \mathcal{X}_{d}
\end{equation}

Since the states and actions can only take discrete integer values, it was necessary to discretise the normal distribution to ensure that transitions between states were always integers. It was decided that the distribution would be centered around 0 and could only take values in the set $\mathcal{X}_{d} \in \{-1,0,1\}$. 
To do this, a continuous normal distribution, described by random variable $\mathcal{X}$, was discretised to a discrete random variable $\mathcal{X}_{d}$.


\subsubsection{Normal Distribution}
To discretise the normal distribution with a set standard deviation, $\sigma$, the probability that the random variable $\mathcal{X}$ will fall between 1 and -1 was calculated. This was computed by using the cumulative distribution function which is defined as $F_{\mathcal{X}}(x) = P(\mathcal{X}\leq x)$:
\begin{equation}
    P(-1<\mathcal{X}\leq1) = F_{\mathcal{X}}(1) - F_{\mathcal{X}}(-1)
\end{equation}

\begin{figure}[H]
    \centering
    \includegraphics[width = 10cm]{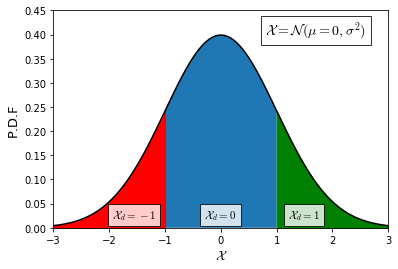}
    \caption{A figure displaying how the normal distribution is discretised. The blue area between -1 and 1 represents the probability of the discrete variable $P(\mathcal{X}_{d} = 0)$. Due to the symmetry of the normal distribution, the area of the red and green regions are equal, representing the fact that $P(\mathcal{X}_{d} = -1) = P(\mathcal{X}_{d} = 1)$. If $\sigma = 1$, $P(\mathcal{X}_{d} = \pm1) \approx 0.16$ and $P(\mathcal{X}_{d} = 0) \approx 0.68$.}
    \label{fig:discretenormal}
\end{figure}

This probability is then assigned to  $P(\mathcal{X}_{d} = 0)$. One then assigns $P(\mathcal{X}_{d} = -1) = P(\mathcal{X}<-1)$ and $P(\mathcal{X}_{d} = 1) = P(\mathcal{X}>1)$. As the normal distribution is symmetric about the mean, these two probabilities are equivalent. This is illustrated in figure \ref{fig:discretenormal}. 

As there is now a chance of the agent going up or down two steps in each timestep, in theory the agent should be able to access a higher state than $S = T$. However a hard boundary is imposed, which means that when the agent takes an action that would take it above or below $S = \pm T$ respectively, it remains at $S = \pm T$.

\section{Results - Sample Trajectories}
The first part of this example was to simply demonstrate the optimisation of the policy. This was done by first calculating the expected return and generating 100 trajectories from the random policy in equation \ref{randompolicy}. The policy was then optimised, the expected return was calculated again and 100 more trajectories were sampled. This process was repeated for both deterministic and noisy environments, with $T=20$. 

\subsubsection{Deterministic Environment}
\begin{figure}[H]
     \centering
     \begin{subfigure}[b]{0.49\textwidth}
         \centering
         \includegraphics[width=\textwidth]{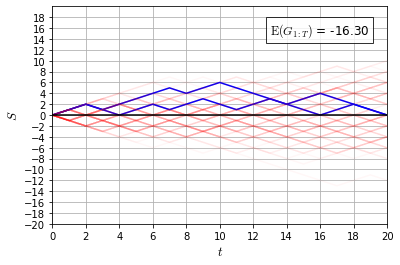}
         \caption{Before Policy Optimisation.}
         \label{fig:y equals x}
     \end{subfigure}
     \begin{subfigure}[b]{0.49\textwidth}
         \centering
         \includegraphics[width=\textwidth]{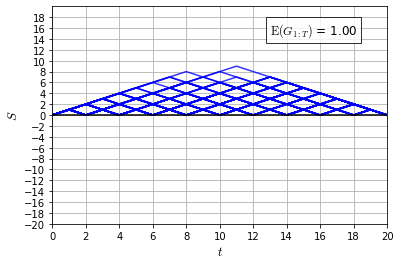}
         \caption{After Policy Optimisation.}
         \label{fig:three sin x}
     \end{subfigure}
     \hfill
        \caption{Showing 100 trajectories that were generated before and after policy optimisation. Red trajectories represent those that did not achieve the maximum return, and blue represents those that did. As there are significantly more red trajectories, they are plotted slightly transparently. Also shown is the expected return. a) There is a large negative expected return, giving rise to a large number of trajectories that do not obey the objective. b) The expected return is now 1, which is the highest possible score achievable in this simulation, and every trajectory obeys the objective.}
        \label{fig:deterministicsarl}
\end{figure}

The deterministic environment represents the situation where there is no noise present. If the agent takes the action to go up, it will go up by exactly one state. 
Figure \ref{fig:deterministicsarl} displays the results of 100 trajectories, plotted before and after policy optimisation. As expected in this deterministic case, policy optimisation is able to find a policy that achieves the best possible expected return, equal to 1. This means that every trajectory generated by the policy will obey the objective perfectly.

\begin{figure}[H]
    \centering
    \includegraphics[width=10cm]{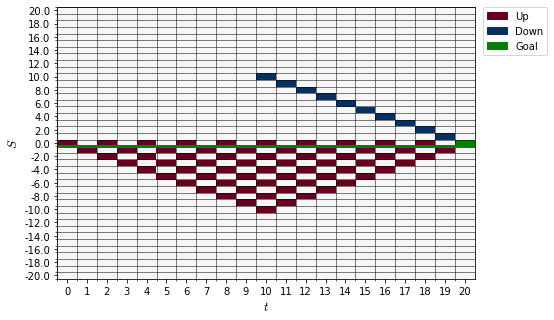}
    \caption{Graphically displaying the policy after optimisation. The `inverse light cone' is clearly shown as the diagonal line of blue squares between $t=\frac{T}{2}$ and $t=T-1$.}
    \label{policy}
\end{figure}

Once the policy has been optimised, it can be displayed graphically, which is shown in figure \ref{policy}. Firstly, there is no preference of action on any state that it is not possible for the agent to land on, for example $S_{t=1} = 0$. Also for any state that is at or below $S=0$, the action is to go up. When above 0, there is no preference of actions until the agent reaches the `inverse light cone', which is a diagonal line of actions to go down, starting at $t = \frac{T}{2}$, which ensures that the agent ends at $S = 0$. 

\subsubsection{Normal Random Noise}
\begin{figure}[H]
     \centering
     \begin{subfigure}[b]{0.49\textwidth}
         \centering
         \includegraphics[width=\textwidth]{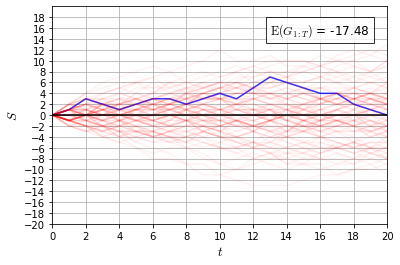}
         \caption{Before Policy Optimisation.}
         \label{fig:y equals x}
     \end{subfigure}
     \begin{subfigure}[b]{0.49\textwidth}
         \centering
         \includegraphics[width=\textwidth]{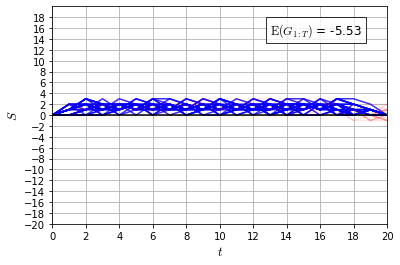}
         \caption{After Policy Optimisation.}
         \label{fig:three sin x}
     \end{subfigure}
     \hfill
        \caption{An analogous diagram to figure \ref{fig:deterministicsarl}, but with the presence of random noise. a) There is a large negative expected return, giving rise to a large number of trajectories that do not obey the objective. These trajectories also appear more diffuse than the deterministic case. b) The expected return can not reach 1, due to the presence of the noise. One can see that the agent stays much closer to 0 than in the deterministic case. In total 30\% of the trajectories obeyed the objective.}
        \label{fig:normalsarl}
\end{figure}

Normally distributed random noise was added to the state transitions in the way described above. This meant that it was now possible to move over more than one state per timestep, or to not move at all. The standard deviation of the normal distribution was set to $\sigma = 1$. 
Figure \ref{fig:normalsarl} shows the results of plotting 100 trajectories before and after policy optimisation. One can see that before policy optimisation, the trajectories are more diffuse that the deterministic case, which is expected as the agent can reach states further away from 0 more quickly, due to its ability to move 2 states per timestep. Also, note that the expected return is lower than in the deterministic case. This is likely due to the fact that there is now more states that the agent can be in at time $T$. Where as before, the agent had to be in an even state (-2,0,2...), now the agent can be in any state. This will increase the probability of  obtaining the large -10 reward for $S_{T} \neq 0$. After policy optimisation we see an interesting trend. The agent learns to stay much closer to 0 than in the deterministic case. Not only this, but the agent also is able to not go below 0 for most of the timesteps. Taking the mean value of the state of the trajectories between times $t=2$ and $t=16$, yields a value of approximately 1.5. Clearly the strategy devised by the agent is to oscillate between $S=1$ and $S=2$. The advantage of this, is going down from 2 can yield a lowest possible state of 0, which does not cause any negative reward. A question remains of why doesn't the agent choose a higher valued state to oscillate around. The answer to this is possibly that when approaching $t=T$, the agent needs to be as close to $S = 0$ as possible. This is because it must account for the fact that when it takes an action to go down, it may not move. Using this strategy inevitably leads to some trajectories that will go below 0, as taking actions to go down from 1 in the final timesteps can cause the agent to go down 2 states to negative states.

\begin{figure}[H]
    \centering
    \includegraphics[width = 9cm]{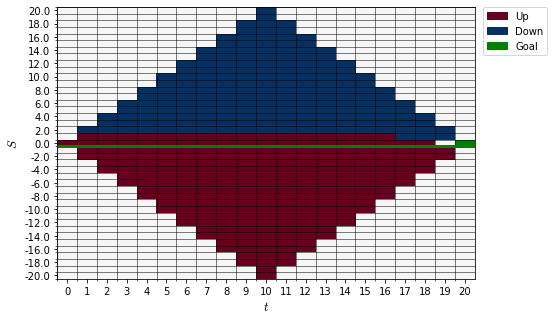}
    \caption{The optimised policy in the case of random noise being present. One can see that for most $t$, the agent oscillates between $S=1$ and $S=2$. }
    \label{normalpolicy}
\end{figure}

The optimal policy is also plotted in figure \ref{normalpolicy}. The reverse light cone is still present, but it now jumps 2 steps at a time, as the agent has the possibility of jumping 2 at once. The other difference is that there is a preferred action for many more states than in the deterministic case. This is simply because it is possible for the agent to be in these states now.

\section{Planning}\label{planning}
When performing reinforcement learning in a practical setting, one often does not know the dynamics of the environment. To learn how to act, one may implement planning, which involves the agent learning a model of its environment, and then optimising its policy with respect to this model \cite{planning}. The agent starts with a model that is completely random, with equal probabilities of obtaining a pair of $s'$ and $r$ given $s$ and $a$ ($P(s',r| s,a)= \frac{1}{N_{S}\times N_{R}} \; \forall \; s',r,s,a$). In practice, this means that every element of the agent's model of the $\mathbf{M}_{t}$ tensor is the same. The agent also starts with an initially random policy and then repeats the following process iteratively until convergence. 
\begin{enumerate}
\item Sample $N_{\textrm{traj}}$ trajectories from the true environment, using its current policy.
\item Use this information to update its model of the environment.
\item Optimise policy with respect to model environment.
\end{enumerate}

One cycle through the steps shown above is defined as one epoch. The first step is to generate trajectories. The agent starts in state $s = 0$. It then finds the action from the policy and uses this to transition to the next state, $s'$, where it obtains a reward $r$. It repeats this process until all timesteps have finished. When doing this process an exploration rate, $\epsilon$, is used so that the agent continues to explore the state space even once the policy has been optimised. This is essential in order to discover better policies than the one already found. To implement the exploration rate, when the agent takes an action from the policy at time $t$, a uniformly random number, $\mathcal{U} \in [0,1]$, is generated. If $\mathcal{U} < \epsilon$ then $A_{t} \xrightarrow{} -A_{t}$. Clearly if $\epsilon = 0$, the agent will follow the policy exactly.

The next step involves the agent updating its model of the $\mathbf{M}_{t}$ tensors, which one labels $_{\textrm{model}}\mathbf{M}_{t}$. It does this based on information it obtained from step 1. In this experiment, we assume that the agent has the knowledge that all $\mathbf{M}_{t}$ tensors are the same, except the one for $t=T$. To update element $_{\textrm{model}}\mathbf{M}_{t}^{\,s',r,s,a}$, which corresponds to a probability $P(s',r|s,a)$, one studies the trajectories and first finds every transition where the $s,a$ pair was accessed. Then one finds the fraction of times where $s',r$ was the outcome of the transition, which one can label as $\textrm{NEW}$. One then updates the value in the $_{\textrm{model}}\mathbf{M}_{t}$ tensor using a convex combination, which ensures that the new values in $_{\textrm{model}}\mathbf{M}_{t}$ still represent probabilities \cite{convexcombination}:

\begin{equation}
    _{\textrm{model}}\mathbf{M}_{t}^{\,s',r,s,a} \xrightarrow[]{}\, _{\textrm{model}}\mathbf{M}_{t}^{\,s',r,s,a} + \alpha \left(\textrm{NEW} - \,_{\textrm{model}}\mathbf{M}_{t}^{\,s',r,s,a}\right)
\end{equation}

Another variable, $\alpha \in [0,1]$, has been introduced which is the learning rate. This determines how much importance to assign to the fraction calculated by the new trajectories, compared to previous steps of trajectory sampling. One repeats the above process for all elements of $_{\textrm{model}}\mathbf{M}_{t}$. Once the model has been updated, the policy is optimised with respect to it, using the DMRG algorithm. At the end of each epoch, the expected return with respect to both the true environment and model environment can be calculated. 

\subsection{Results - Planning}

\subsubsection{Deterministic Environment}
\begin{figure}[H]
    \centering
    \includegraphics[width = 11cm]{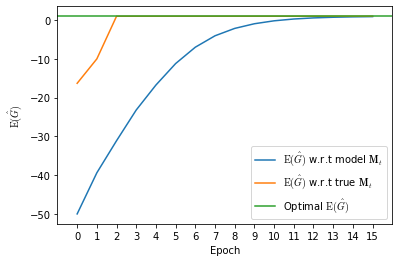}
    \caption{An example graph showing the effects of planning. To generate this plot the following parameters were used, $\alpha = 0.4$, $\epsilon = 0.2$ and $N_{\textrm{traj}} = 30$. }
    \label{fig:deterministicsarlplan}
\end{figure}

Figure \ref{fig:deterministicsarlplan} displays the results of implementing planning in a deterministic environment. 
At the first epoch, the expected return with respect to the model starts at $\mathbb{E}(G) = -50$. The reason for this is that there is initially an equal chance of obtaining any of the rewards in each timestep so that the average reward per timestep is the mean of the possible rewards. The possible rewards are $[1,0,-1,-10]$, which have a mean value of -2.5. Multiplying this by the number of timesteps (20) yields the initial expected return of -50.

The expected return with respect to the true environment increases  to -10 in the first epoch. When I investigated the optimised policy after the first epoch it was clear that the agent learned to stay above 0, but did not learn to end on 0 at $t=T$, which is why its $\mathbb{E}(\hat{G}) = -10$. After generating another 30 trajectories, the agent learned to end on $S=0$, giving it the optimal return. The expected return with respect to the model environment increases gradually to the optimal value. This is because its state transitions are made more accurate in each timestep. 

It was found that the expected return with respect to the true environment converged to the optimum value very quickly. If $N_{\textrm{traj}} > 10$, it would consistently converge in two epochs, and often in one.  

\subsubsection{Noisy Environment}
\begin{figure}[H]
    \centering
    \includegraphics[width = 11cm]{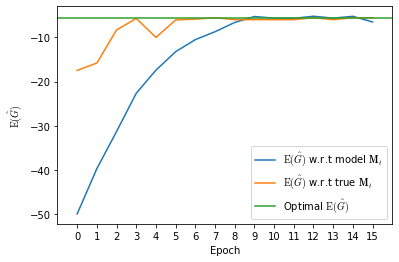}
    \caption{An example graph showing the effects of planning, in the noisy environment. To generate this plot the following parameters were used, $\alpha = 0.4$, $\epsilon = 0.2$, $N_{\textrm{traj}} = 30$ and $\sigma = 1$. }
    \label{fig:normalsarlplan}
\end{figure}

In the case of the noisy environment, it is expected that it should be more difficult for the agent to learn the state transitions in its model, due to the fact that there are multiple possibilities for the next state when taking a set action. Therefore, in theory one would expect that $\mathbb{E}(\hat{G})$ w.r.t the model would lag further behind the $\mathbb{E}(\hat{G})$ w.r.t the true environment than in the deterministic case. However figure \ref{fig:normalsarlplan} shows that $\mathbb{E}(\hat{G})$ w.r.t the model approaches the optimal value after approximately 9 epochs, whereas it took 12 in the deterministic case. A possible reason for this is that the noise meant that more of the state space was explored across the trajectories. In other words, in the deterministic case, the same trajectories were being generated fairly regularly. In this case however, the noise meant that a wider range of trajectories were generated, and so more state transitions were able to be learnt.

One also sees that after both expected returns have reached the optimal value, they fluctuate around it, due to the state transitions for the model being continuously updated. Consider the state transition $s_{T-1} = 1 \xrightarrow[]{} s_{T} = -1$. During an epoch of trajectory generation, it is statistically possible for a large quantity of these transitions to be generated. In this case, the probability of this occurring in the model will be increased, which in turn could cause the policy to be optimised in a way that decreases $\mathbb{E}(\hat{G})$ w.r.t the true environment. This signifies the importance of not making the learning rate, $\alpha$, too high. By increasing the significance of previous trajectories, one can obtain better approximations of the state transitions, as the averages are being taken over more data. Of course, the number of trajectories can also have a significant effect on how much the expected returns fluctuate. For a very high $N_{\textrm{traj}}$, much of the state space will be explored in each epoch, and thus the state transitions will be learnt more accurately. 
\chapter{Multi-Agent RL with TNs}

\ifpdf
    \graphicspath{{Chapter5/Figs/Raster/}{Chapter5/Figs/PDF/}{Chapter5/Figs/}}
\else
    \graphicspath{{Chapter5/Figs/Vector/}{Chapter5/Figs/}}
\fi
\section{Multi-Agent Reinforcement Learning}
Multi-agent reinforcement learning (MARL) is the analogue of single-agent RL (SARL), in the case when more than one agent is being considered. The agents share an environment, and each must make a decision every timestep. The evolution of the states and rewards between timesteps, is affected by the actions of all the agents. The majority of algorithms can be placed into three discrete categories, being cooperative learning, competitive learning and then a mix of the two \cite{MARL, MARL2}. In competitive learning, each agent acts to maximise its own score, regardless of the score of the other agent. Reward structures in this setting are often created so that the sum over all of the agent's returns equals 0. This simulates an environment in which there is a finite amount of resources with different agents fighting for them.

This report focuses on cooperative learning, in which agents must work together to achieve a common goal. Here, the assessment of performance is a return composed of the sum of the agent's returns \cite{cooperative, MARL3}. The process displayed in Chapter 3 will be repeated. Firstly a TN will be constructed that has the capability of describing state transitions. Then the rewards will be introduced, before the actions and policy are added at the end. In explanations, a TN will be formulated for the case of two agents, although the construction can be generalised to more. The construction of the MARL TN is highly analagous to the SARL case, and under the assumption that previous chapters have been read, some descriptions will remain brief. 

A TN consisting of joint probability distributions will be introduced first, as this represents the most general case with no assumptions made (except the assumption of a FMDP from the previous sections). However, this structure suffers from the curse of dimensionality when increasing the number of agents. For this reason, the joint distributions will be decomposed.

\section{Tensor Network for State Transitions}
As in the SARL case, the initial probability tensor must be considered first. In the single agent case, there was one probability vector, $\ket{{p}_{0}}$, which represented the probabilities: $P(S_{0} = s)$. In the two agent case, a joint probability is considered:

\begin{equation}
    P(s^{(1)}_{0},s^{(2)}_{0}) = P(S^{(1)}_{0}=s^{(1)}_{0},S^{(2)}_{0}=s^{(2)}_{0})
    \label{jointdist}
\end{equation}

Where the (i) superscript indicates the state of the i\textsuperscript{th} agent. To represent the distribution shown in equation \ref{jointdist} in TN formulation, a matrix, $\ket{{p}^{J}_{0}}$, can be used. 

\begin{equation}
    \includegraphics[width = 6cm]{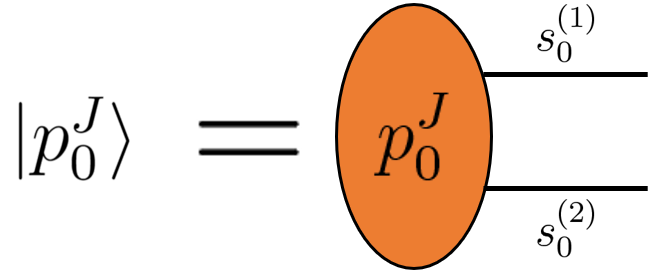}
\end{equation}

One must also adapt the $\mathbf{M}_{t}$ tensors from the SARL explanation to this MARL setting. When considering state transitions without actions or rewards, the relevant probabilities in the single agent case were $P(S_{t}=s_{t}|S_{t-1} = s_{t-1})$. Now two agents are being considered and in the most general case, it is assumed that each of these agents affect the state transitions for the other agent. For this reason, we cannot assume two independent conditional probabilities for each agent, but instead a joint conditional probability. 
\begin{equation}
    P(s^{(1)}_{t},s^{(2)}_{t}|s^{(1)}_{t-1},s^{(2)}_{t-1}) = P(S^{(1)}_{t}=s^{(1)}_{t},S^{(2)}_{t}=s^{(2)}_{t}|S^{(1)}_{t-1} = s^{(1)}_{t-1},S^{(2)}_{t-1} = s^{(2)}_{t-1})
    \label{statetransitionmulti}
\end{equation}
Equation \ref{statetransitionmulti} describes a relation between four components in vector space, so a rank 4 tensor is suitable for describing these transition probabilities. Equation \ref{Mnoartogether} displays the way that this tensor, $\mathbf{M}_{1}^{J}$,  can propagate the probability state vector through time. In the graphical notation, it is shaded in a dark grey once more to represent the fact that it describes transitions between states only.

\begin{equation}
\includegraphics[width=8cm]{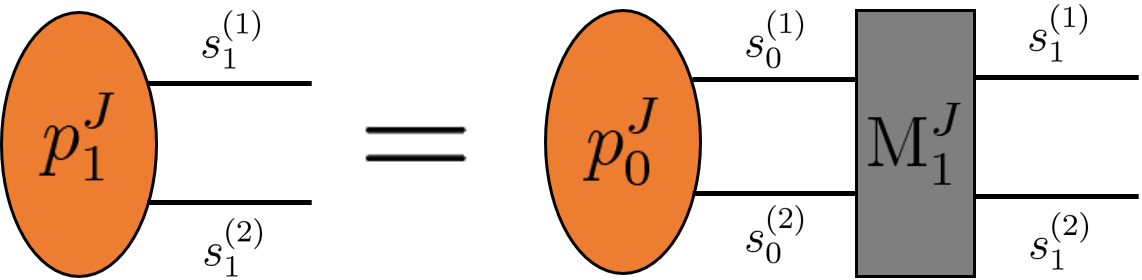}
\label{Mnoartogether}
\end{equation}

One can easily contract a tensor network using these tensors, in a way that is analogous to the single agent case shown in equation \ref{flattensor}. This is shown in equation \ref{joint} for $T=3$.

\begin{equation}
\includegraphics[width=12cm]{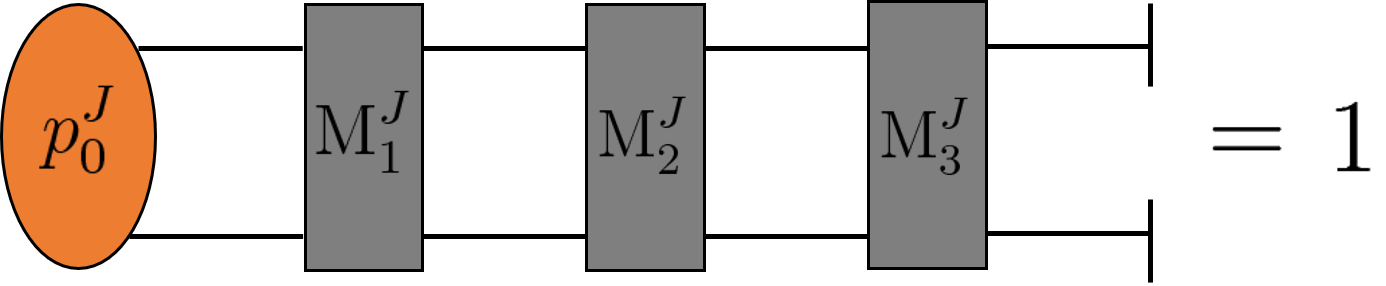}
\label{joint}
\end{equation}

\section{Tensor Network with Rewards}

To accommodate for rewards, the conditional probability represented by $\mathbf{M}_{t}^{J}$ is updated to $P(s_{t}^{(1)},s_{t}^{(2)},r_{t}^{(1)},r_{t}^{(2)}|s_{t-1}^{(1)},s_{t-1}^{(2)})$. Therefore, the rank of the $\mathbf{M}_{t}^{J}$ tensor increases by two, so that it is now rank 6. To represent the $r^{(i)}_{t}$ indices using the graphical notation, one marks them as red diagonal lines, representing that they are going into the page. As before, this tensor is related to the previous form of $\mathbf{M}_{t}^{J}$ (that just represented state transitions) by marginalisation, which is shown by contracting each of the reward indices with flat reward tensors.
\begin{equation}
    \includegraphics[width = 9cm]{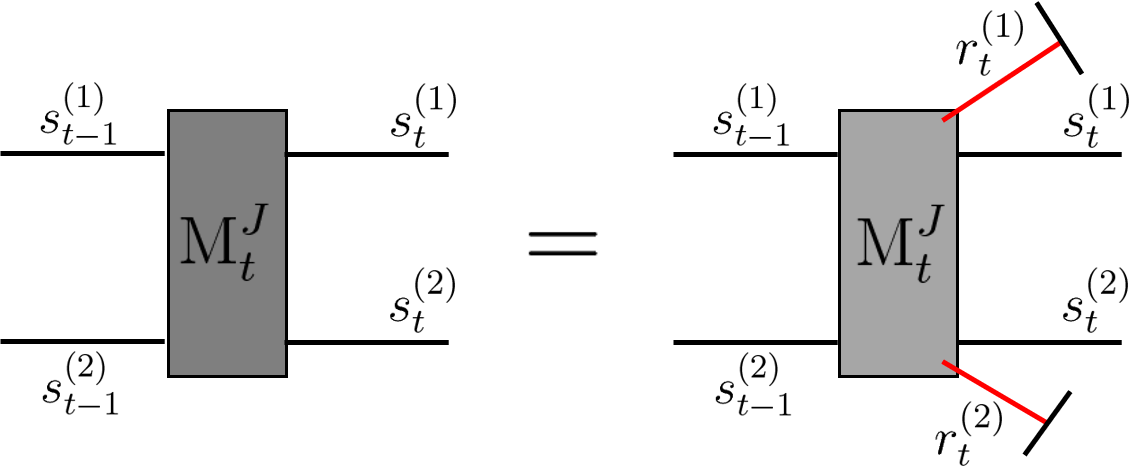}
\end{equation}

As previously discussed, the model being explained is for cooperative learning, in which the total return comes from the sum of rewards over agents and timesteps.
\begin{equation}
    G_{1:T} = \sum_{t=1}^{T}\sum_{i = 1}^{2}r^{(i)}_{t}
\end{equation}
One must now find suitable tensors, $\Tilde{\mathbf{W}}^{J}_{t}$, so the rewards are both summed between agents, and across all timesteps. In the case of rewards, it is actually easier to find the form of the $\Tilde{\mathbf{W}}^{(i)}_{t}$ tensors, where there exists one of these tensors for each agent per timestep.

One can exploit the fact that the sum between agents can be considered in the same way as the sum between timesteps. The exact same $\Tilde{\mathbf{W}}^{(i)}$ tensors from the single agent case can be used, arranged in a snaking shape which is shown for $T=3$ on the left hand side of equation \ref{wlayer}. In the first timestep, one can treat agent 1's reward as being summed before agent 2's. Then in the second timestep agent 2's is summed before agent 1's. The first and second timesteps are summed by contracting the tensors corresponding to agent 2 at $t=1$ and $t=2$. This process can be repeated for all timesteps, which results in the sum of all rewards for all agents and timestep. The resulting tensor network consists of $2\times T$ $\Tilde{\mathbf{W}}^{(i)}$ tensors.

 Importantly the only tensors that are rank 2 (vectors of vectors) are the first and last ones in the snaking shape, while the remainder are rank 3. In the example explained above, this means that $\mathbf{\tilde{w}}_{1}^{(1)}$ is rank 2, while $\mathbf{\tilde{W}}_{1}^{(2)}$ is rank 3.

\begin{equation}
    \includegraphics[width = 12cm]{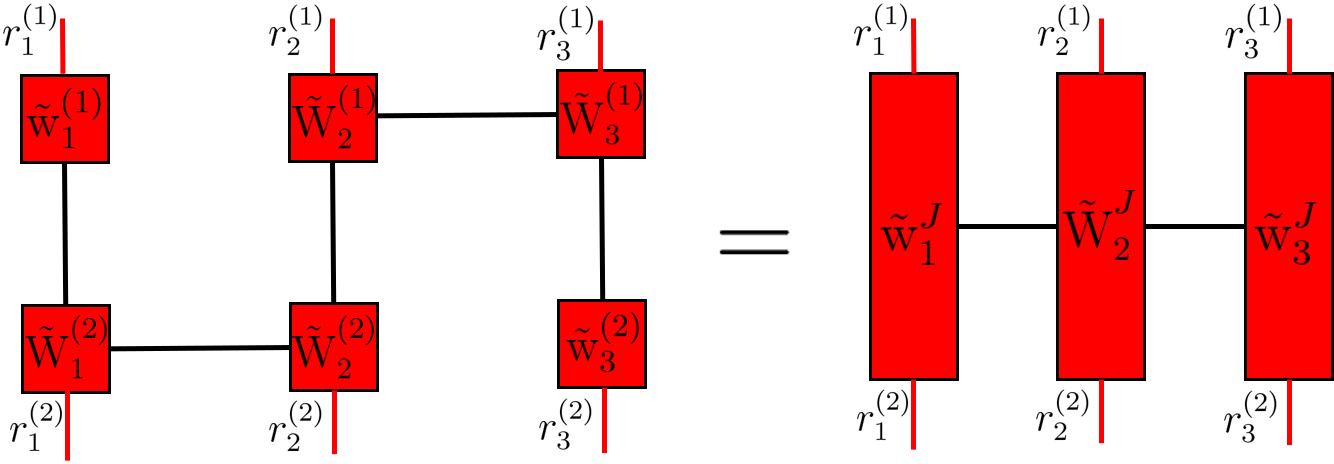}
    \label{wlayer}
\end{equation}

If one wished to have one reward tensor per timestep, they would simply contract the tensors together for all agents per timestep. This is shown on the right hand side of equation \ref{wlayer}. The construction of these tensors is displayed below.

\begin{equation}
    \Tilde{\mathbf{W}}_{t}^{J} = \begin{bmatrix}
\vec{1} & \vec{0}\\
\vec{R} & \vec{1}
\end{bmatrix}\begin{bmatrix}
\vec{1} & \vec{0}\\
\vec{R} & \vec{1}
\end{bmatrix} = \begin{bmatrix}
\vec{1}\otimes\vec{1} & \vec{0}\otimes\vec{0}\\
\vec{R}\otimes\vec{1} + \vec{1}\otimes\vec{R} \,& \vec{1}\otimes\vec{1}
\end{bmatrix}
\end{equation}

\begin{equation}
    \Tilde{\mathbf{w}}_{1}^{J} = \begin{bmatrix}
\vec{R} & \vec{I}
\end{bmatrix}\begin{bmatrix}
\vec{1} & \vec{0}\\
\vec{R} & \vec{1}
\end{bmatrix} = 
\begin{bmatrix}
\vec{R}\otimes\vec{1} + \vec{1}\otimes\vec{R} & \vec{1}\otimes\vec{1} 
\end{bmatrix}
\end{equation}

\begin{equation}
    \Tilde{\mathbf{w}}_{T}^{J} = \begin{bmatrix}
\vec{1} & \vec{0}\\
\vec{R} & \vec{1}
\end{bmatrix} 
\begin{bmatrix}
\vec{1}\\
\vec{R}
\end{bmatrix}
= 
\begin{bmatrix}
\vec{1}\otimes\vec{1} \\
\vec{R}\otimes\vec{1} + \vec{1}\otimes\vec{R}
\end{bmatrix}
\end{equation}
As the tensor product has been used, it is obvious that $\Tilde{\mathbf{W}}_{t}^{J}$ is now a matrix of matrices, while $\Tilde{\mathbf{w}}_{1}^{J}$ and $\Tilde{\mathbf{w}}_{T}^{J}$ are vectors of matrices.

In the remainder of this report, when one sees the red lines representing the reward index, one can assume that it is being contracted with the layer of $\mathbf{\Tilde{W}}^{J}$ tensors described above. One final point on the rewards, is that this structure is very easily generalised to more agents, and more time periods. For example, if one was working with four agents instead of two and $T=4$, the TN of $\mathbf{\Tilde{W}}$ tensors would have the form shown below. On top of this, the computational time required to contract these tensors scales linearly with the numbers of timesteps and agents, as the ranks of the individual tensors do not increase.

\begin{equation}
    \includegraphics[width = 7cm]{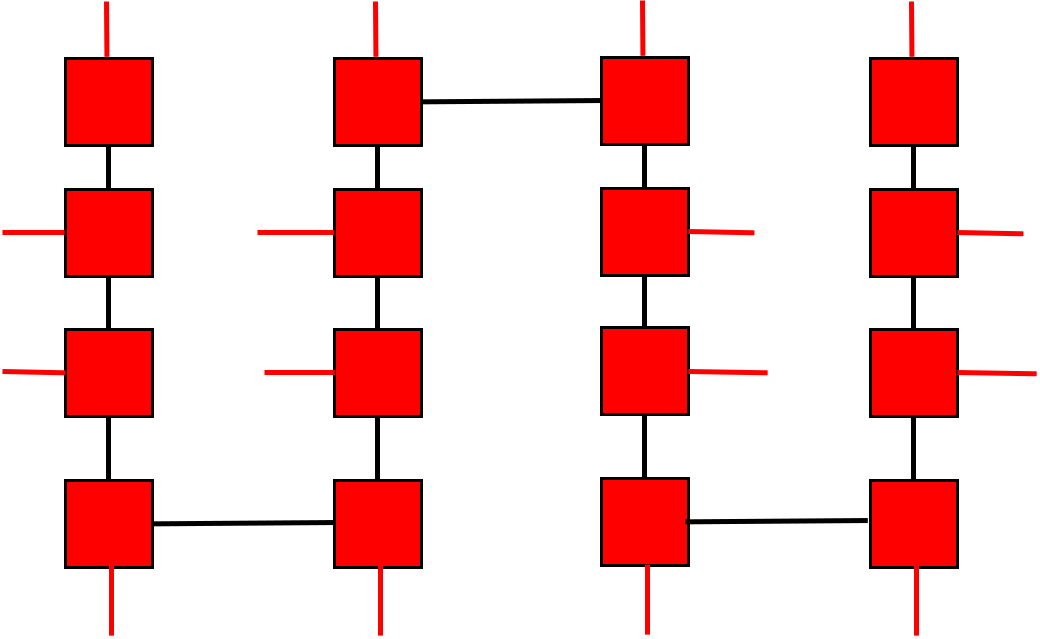}
    \label{4reward}
\end{equation}

\section{Tensor Network with Actions}
The final things to add are the actions for each agent. According to general MARL formalism, each agent's action depends on all of the agent's previous states. This information can be stored in a joint policy, representing a joint probability of all the agent's actions, conditioned on all the agent's previous states. In the two agent example that has been followed so far, the joint policy has the form: $\bm{\pi}_{t}^{J} =  P(a_{t-1}^{(1)},a_{t-1}^{(2)}|s_{t-1}^{(1)},s_{t-1}^{(2)})$. To send the information about the states to each policy, copy tensors analagous to the one used in the SARL case can be implemented. As two indices are being copied, a pair of copy tensors are required. To simplify graphical notation, the pair of copy tensors will be represented by one single black dot:
\begin{equation}
    \includegraphics[width = 8cm]{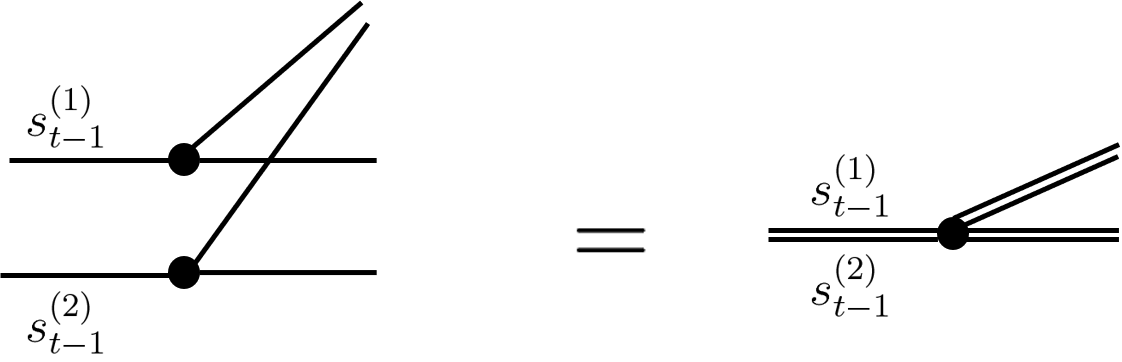}
\end{equation}

Apart from introducing the policy, the only other change is to increase the rank of the $\mathbf{M}_{t}^{J}$ by 2 to account for the actions. The probability represented by $\mathbf{M}_{t}^{J}$ is now $P(s_{t}^{(1)},s_{t}^{(2)},r_{t}^{(1)},r_{t}^{(2)}|s_{t-1}^{(1)},s_{t-1}^{(2)},a_{t-1}^{(1)},a_{t-1}^{(2)})$. The way that the $\mathbf{M}^{J}_{t}$ tensor is contracted with $\pi_{t}^{J}$  is shown below, in an equation that relates this tensor to the previous form of $\mathbf{M}_{t}^{J}$:

\begin{equation}
    \includegraphics[width= 12cm]{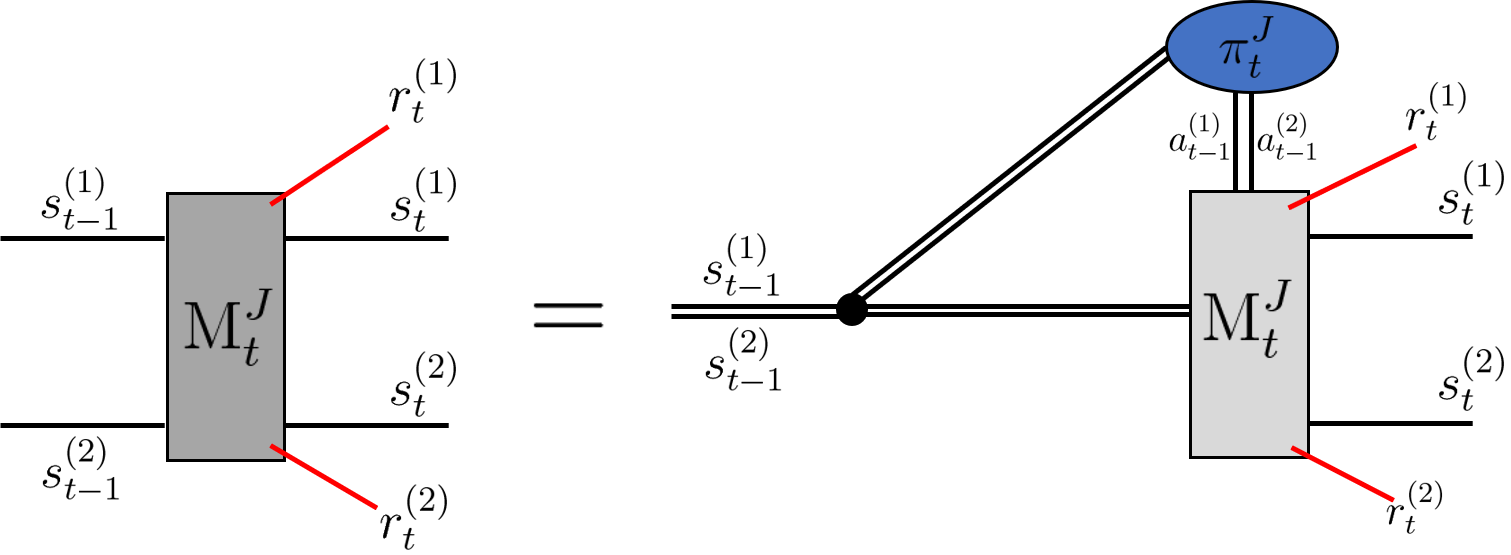}\label{marlpolicy}
\end{equation}

These objects can be contracted together to give the TN representation of the expected return of a FMDP in the MARL setting. An example of this is shown for the case $T=3$ in equation \ref{marlfinal}. It is imperative to remember that the red lines corresponding to the $r_{t}^{(i)}$ indices are being contracted with the layer of $\mathbf{W}_{t}^{J}$ tensors from equation \ref{wlayer}:

\begin{equation}
    \includegraphics[width = 14cm]{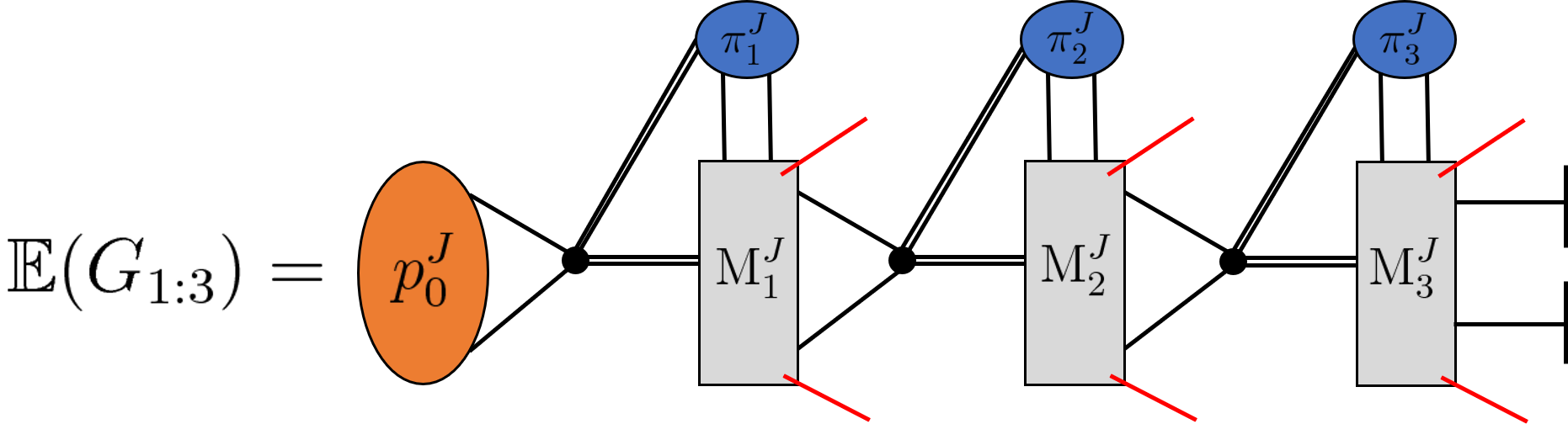}\label{marlfinal}
\end{equation}

The issue with the TN shown in equation \ref{marlfinal}, is that it is subject to the curse of dimensionality. For every extra agent added, the rank of the $\mathbf{M}_{t}^{J}$ tensors will increase by 4. As previously mentioned, this will cause an exponential increase in the number of values being stored when one tries to create this tensor, and consequently simulations with a large amount of agents will not be physically possible to run.  Furthermore, the rank of other tensors such as $\mathbf{p}_{0}^{J}$ and $\bm{\pi}_{t}^{J}$ will also increase linearly with rank. When this problem was encountered in the time dimension, the assumption of a FMDP was used to ensure the number of elements in the $\mathbf{M}_{t}$ tensors increased linearly with time. To minimise the effect of the curse of dimensionality in the agent dimension, one can first decompose the joint tensors so that there is one tensor per agent per timestep. Whilst doing this, one can make approximations that decrease the total number of elements in the tensors by a large amount, making calculations more possible. In the proceeding discussion, the decompositions will first be displayed on the initial probability matrix $\ket{{p}_{0}^{J}}$ for simplicity. 
\section{Decomposition of Joint Tensors}\label{decompositionsection}
There are many possible ways to decompose the tensors. One of which is a natural use of the chain rule, where a joint probability can be broken up into the product of a conditional probability and a singular probability \cite{introprob}. For example, imagine a case where it was known that the first state of each agent were independent of each other. Recall that $\ket{{p}_{0}^{J}}$ represents $P(S_{0}^{(1)},S_{0}^{(2)})$, and in this case it can be decomposed in the following way:

\begin{equation}
    P(S^{(1)}_{0},S^{(2)}_{0}) = P(S^{(1)}_{0}|S^{(2)}_{0})P(S^{(2)}_{0}) = P(S^{(1)}_{0})P(S^{(2)}_{0})
\end{equation}

One can see that a tensor that originally had $({N_{S}})^{2}$ elements can now be written in terms of two tensors with a combined total of $2N_{S}$ elements.
The main advantage of using the chain rule is that all resulting tensors continue to represent probability distributions. However, one issue with decomposing joint distributions in this way is that there is a potential loss of symmetry. The result is that in each timestep, one of the agents essentially has operations performed on it before the other, before the information is passed to the other agent. Moreover, in a lot of situations the assumption that the first states were independent may not be known. It is helpful to have a technique that does not depend on the users knowledge.

If one wishes to maintain symmetry throughout the TN, a technique from linear algebra can be used, singular value decomposition (SVD) \cite{SVD1}. As symmetry means that neither agent is considered before the other, this can be seen as a better representation of the general case than when using the chain rule. SVD can also be considered more flexible, as one can choose the level of decomposition. For these reasons, SVD will be used in the graphical notation for the remainder of the report.
\subsubsection{Singular Value Decomposition}
SVD can be used to decompose a matrix with indices $s$ and $s'$ into two matrices that each have only one of the indices $s$ or $s'$ \cite{SVD2}. The other index is one that is shared by both matrices representing the correlation between $s$ and $s'$.
Consider a matrix, $\mathbf{p}$ with indices $s$ and $s'$, in analogy to $\ket{{p}_{0}^{J}}$ with indices $s_{0}^{\,(1)}$ and $s_{0}^{\,(2)}$. Each of the indices $s$ and $s'$ may take $N_{S}$ possible values. SVD involves factorising this matrix into three other matrices, $\mathbf{U},\mathbf{V}$ and $\lambda$,

\begin{equation}
    \mathbf{p}^{s,s'} = \sum_{i = 1}^{\chi} \mathbf{U}^{s,i}\bm{\lambda}^{i,i}\mathbf{V}^{i,s'}
    \label{schmidt}
\end{equation}
$\mathbf{U}$ and $\mathbf{V}$ are both unitary matrices and $\bm{\lambda}$ is a diagonal matrix with elements in descending order \cite{edbook}. $\chi$ is known as the Schmidt rank, and the magnitude of it can be thought of as the level of dependence between the indices $s$ and $s'$. For example, the value of $\chi$ must be within the range $1\leq \chi \leq N_{S}$. In the case of $\chi = N_{S}$, every single value of $s$ depends on every value of $s'$. These dependencies are stored in the Schmidt coefficients matrix, $\bm{\lambda}$. At the other extreme, if $\chi = 1$, the indices $s$ and $s'$ are totally independent which can be seen below in equations \ref{independant} and \ref{independantprob}, which show the relation in terms of tensors and probabilities respectively:

\begin{equation}
    \mathbf{p}^{s,s'} =  \mathbf{U}^{s}\mathbf{V}^{s'}
    \label{independant}
\end{equation}
\begin{equation}
    P(s,s') =  P(s)P(s')
    \label{independantprob}
\end{equation}

The general case shown in equation \ref{schmidt} can easily be expressed in graphical notation. This notation can be further simplified by contracting $\bm{\lambda}$ with $\mathbf{V}$ to create $\mathbf{\Tilde{V}}$. This is shown in equation \ref{svd}. One further important point on SVD is that it is a linear algebra method, rather than a probability based method. This means the resulting elements of the tensors, $\mathbf{U}$ and $\mathbf{\tilde{V}}$ no longer represent probabilities. The index $i$ is commonly referred to as the bond dimension:

\begin{equation}
    \includegraphics[width = 9cm]{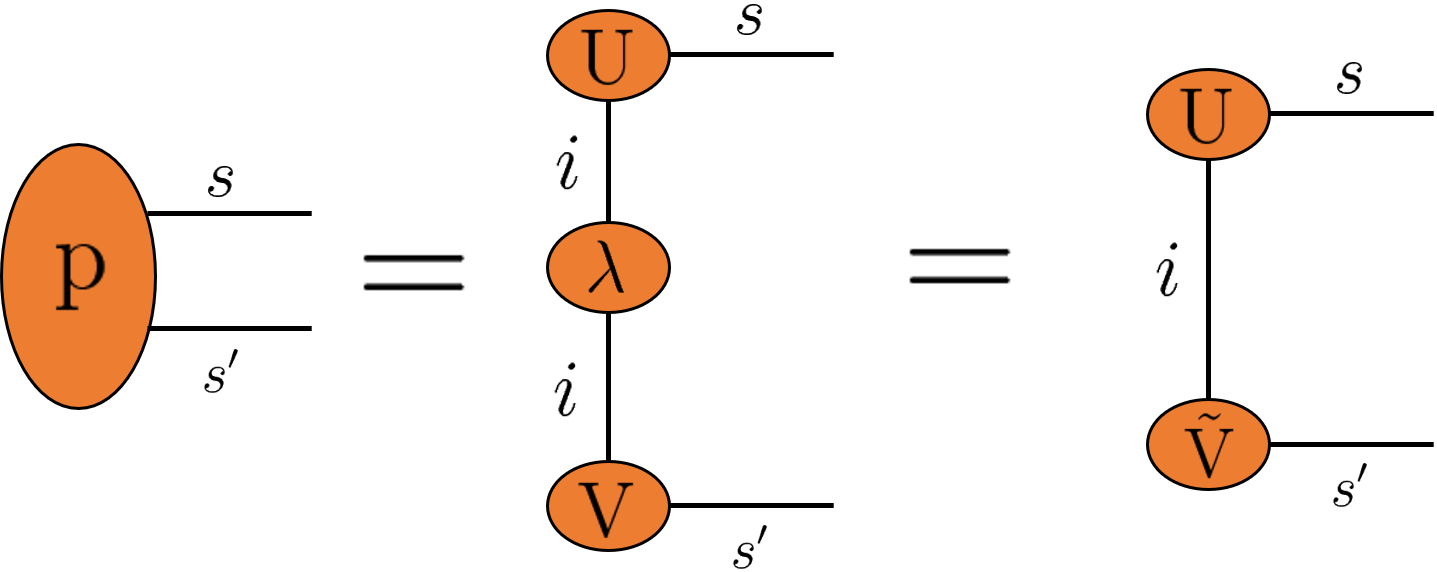}\label{svd}
\end{equation}

There is a value of $\chi$ which causes equation \ref{schmidt} to be satisfied. However, one can deliberately choose a value for $\chi$ that is lower than this value, in order to decrease the size of $\mathbf{U}$ and $\mathbf{\tilde{V}}$. In doing this, one effectively decreases the length of the index $i$ in equation \ref{svd}. This can be used to easily control the compression of the tensors, and to find a lower order approximation. This is particularly useful in situations where the dependence is largely redundant, or where the original tensors are too large to be operated on computationally. One can simply choose a value $\chi$ that allows the calculation to be possible on the machine they're using.

The application of this method to the joint probability matrix $\ket{{p}_{0}^{J}}$ is shown in equation \ref{marlinitialprobvector}, with the shared index labelled $\chi$ to indicate SVD has been used. An arrow has been used instead of an equals sign, as there exists the case of the value of $\chi$ being restricted to find a lower order representation (in which case the equality is lost):

\begin{equation}
    \includegraphics[width = 8cm]{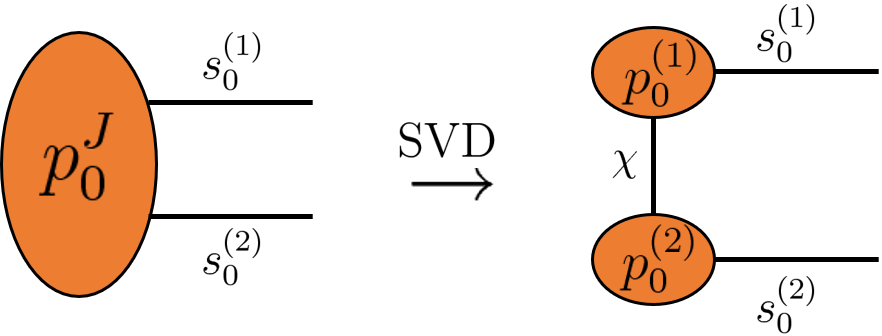}\label{marlinitialprobvector}
\end{equation}
To generalise this structure to the case of more agents, one would sort the indices of the agents together into two groups, before performing SVD. Then one can perform SVD again to further split up the agents. This is shown explicitly for the case of four agents in equation \ref{decompose4}:

\begin{equation}
    \includegraphics[width = 13cm]{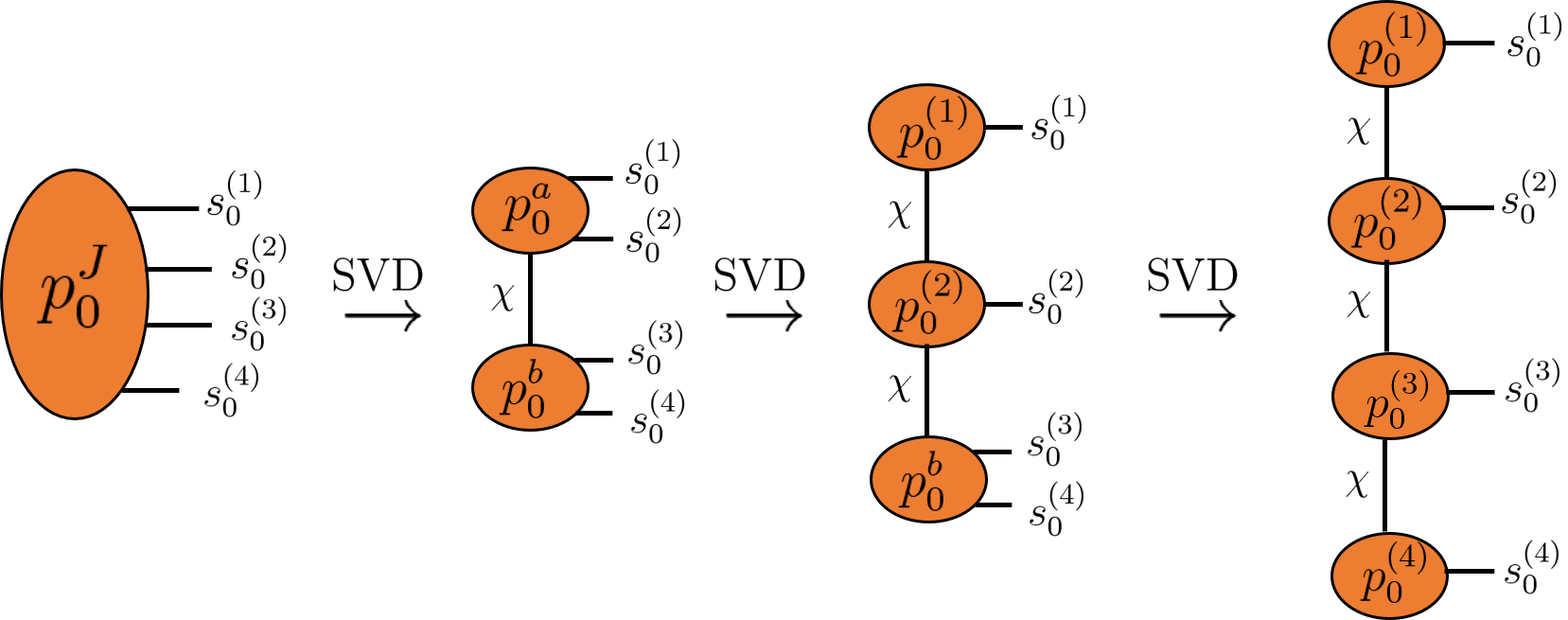}\label{decompose4}
\end{equation}

Of course one also wishes to decompose the rank 8 $\mathbf{M}^{J}_{t}$ tensors in equation \ref{marlfinal}. After all, it is these tensors that will have the most severe increase to their rank when an extra agent is added. To get these tensors into the form that is applicable to equation \ref{schmidt}, one must flatten $\mathbf{M}^{J}_{t}$ into a rank 2 tensor (a matrix). It is natural to flatten the eight indices so that the four indices of agent 1 ($S_{t}^{(1)},R_{t}^{(1)},S_{t-1}^{(1)},A_{t-1}^{(1)}$) are combined into one index in the flattened tensor, and the four indices of agent 2 are combined into the other. This is the logical choice, as one is trying to find the dependence between the agents after all. One can then perform SVD to find the two tensors $\mathbf{M}^{(1)}_{t}$ and $\mathbf{M}^{(2)}_{t}$. Finally the tensors can be unfolded so that they have the indices $S_{t}^{(i)},R_{t}^{(i)},S_{t-1}^{(i)},A_{t-1}^{(i)}$ and the $\chi$ index from the SVD. This method is shown in figure \ref{fig:svdm}:

\begin{figure}[H]
    \centering
    \includegraphics[width = 14cm]{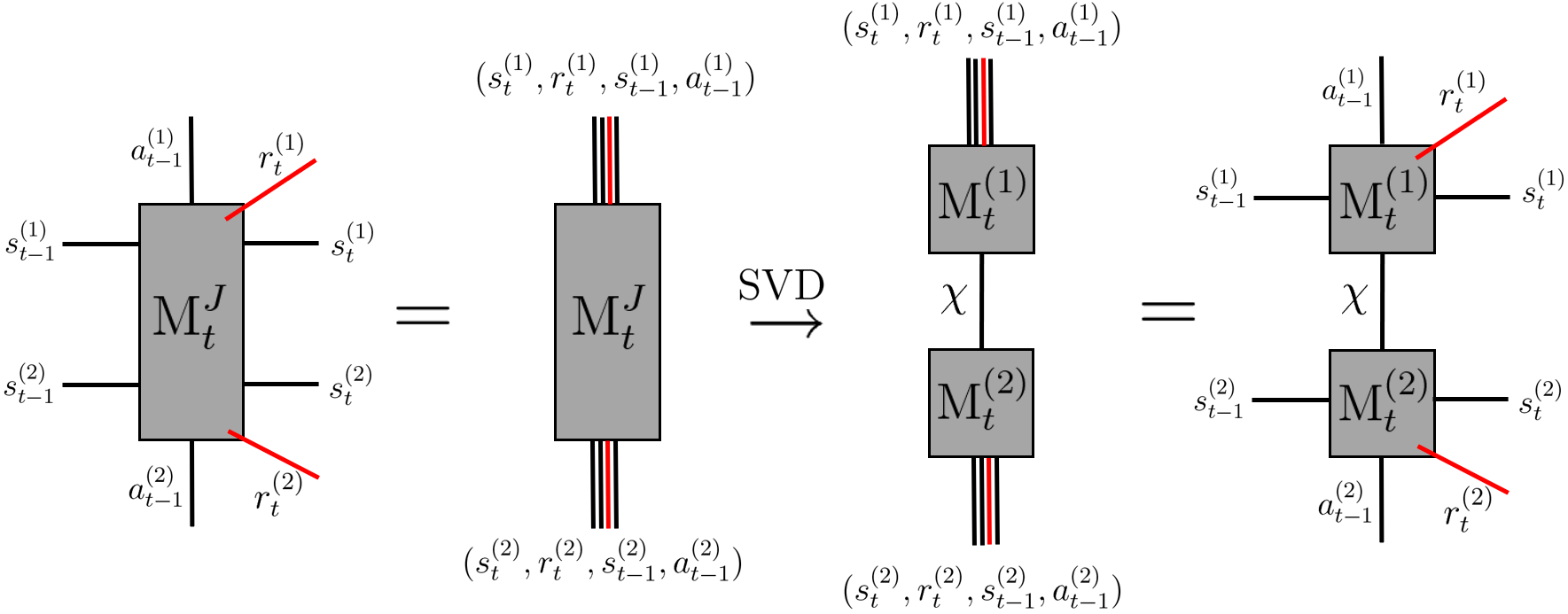}
    \caption{Showing the process of decomposing $\mathbf{M}_{t}^{J}$ into two tensors $\mathbf{M}^{(1)}_{t}$ and $\mathbf{M}^{(2)}_{t}$. First the indices of the tensor are grouped into two indices. Then SVD is performed to decompose into two matrices. Finally, these are unflattened. The application of this decomposition is shown in section 6 of this report.  }
    \label{fig:svdm}
\end{figure}

 The other tensor that currently grows exponentially with increasing agents is the policy, $\bm{\pi}^{J}_{t}$. To decompose this, one repeats the process that was applied to the $\mathbf{M}_{t}^{J}$, beginning with grouping the indices of each agent into two large indices. The resulting tensors, $\bm{\pi}^{(1)}_{t}$ and $\bm{\pi}^{(2)}_{t}$, are shown in equation \ref{decompose policy}:

\begin{equation}
    \includegraphics[width = 10cm]{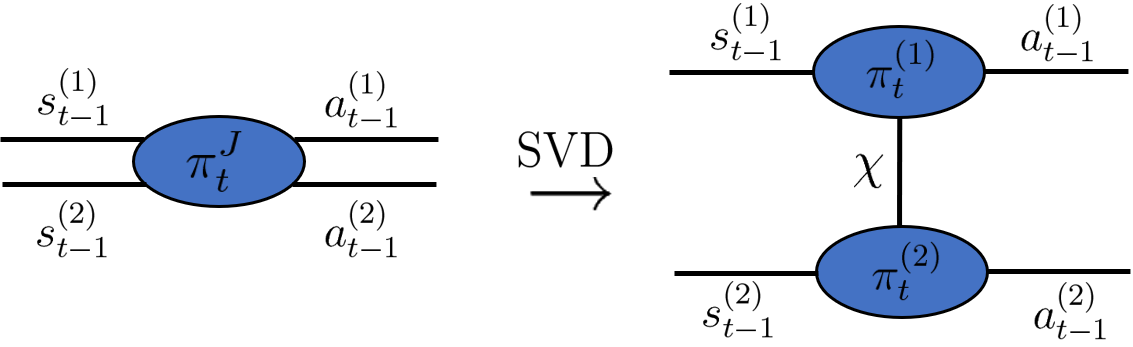}\label{decompose policy}
\end{equation}

\begin{figure}[H]
     \centering
     \begin{subfigure}[b]{0.45\textwidth}
         \centering
         \includegraphics[width=\textwidth]{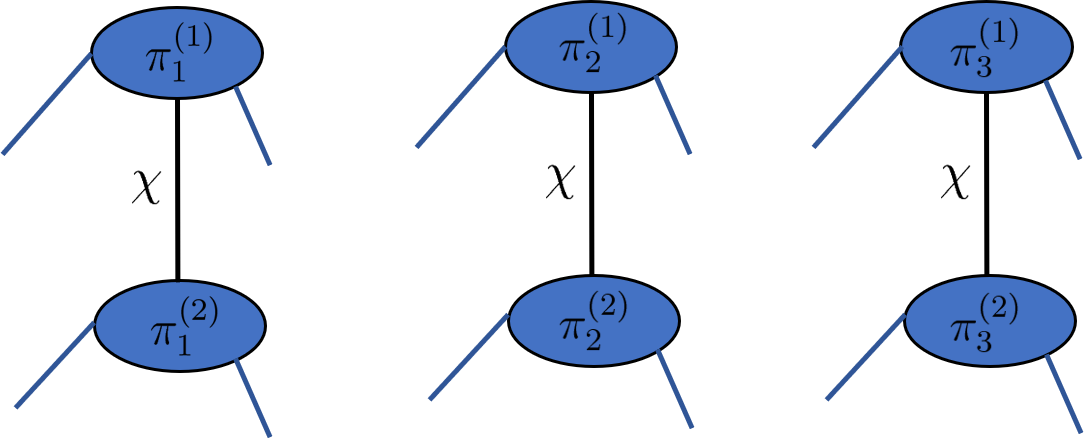}
         \caption{Top Layer}
         \label{fig:y equals x}
     \end{subfigure}
     \hfill
     \begin{subfigure}[b]{0.45\textwidth}
         \centering
         \includegraphics[width=\textwidth]{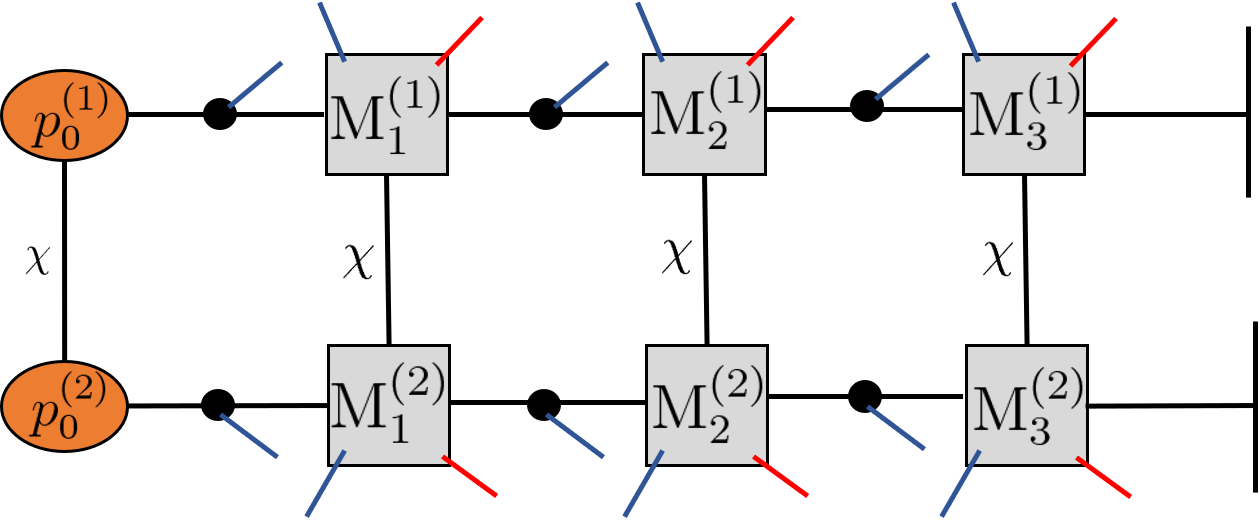}
         \caption{Middle Layer}
         \label{fig:three sin x}
     \end{subfigure}
     \begin{subfigure}[b]{0.45\textwidth}
         \centering
         \includegraphics[width=\textwidth]{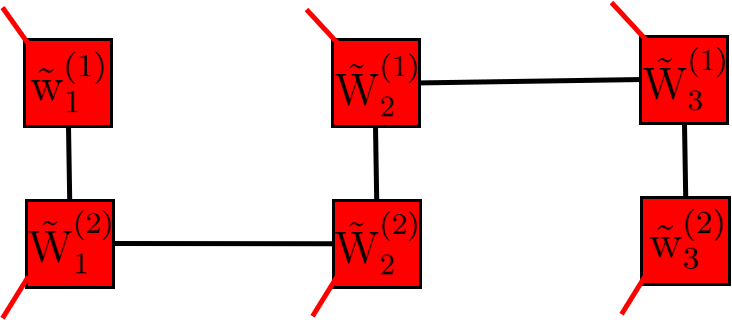}
         \caption{Bottom Layer}
         \label{fig:five over x}
     \end{subfigure}
     \hfill
      \begin{subfigure}[b]{0.45\textwidth}
         \centering
         \includegraphics[width=\textwidth]{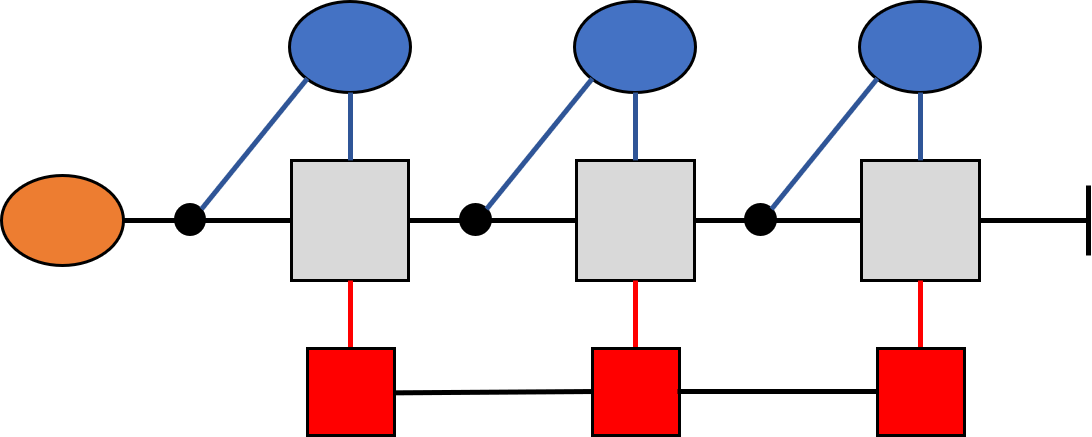}
         \caption{Side On}
         \label{fig:five over x}
     \end{subfigure}
        \caption{Displaying the tensor network with the joint tensors decomposed. a), b) and c) display diagrams of the top, middle and bottom layers respectively. d) displays a side on perspective. Black lines represent contractions within a layer. Blue lines represent contractions between the middle and top layers and red lines represent contractions between the middle layer and the bottom layer.}
        \label{decomposemarlfinal}
\end{figure}

These tensors can be contracted with the network using copy tensors. It becomes difficult to display the tensor network at this point, as it has a 3D structure. It consists of a 'middle' layer of $\mathbf{M}_{t}^{(i)}$ tensors, sandwiched between a 'top' layer of $\bm{\pi}_{t}^{(i)}$ tensors and a 'bottom' layer of $\mathbf{W}_{t}^{(i)}$ tensors. This structure is shown in figure \ref{decomposemarlfinal}. $\chi$ is used to denote where SVD has been used, but it's important to note that the length of the bond dimension will not be the same in the case of the $\pi_{t}^{(i)}$ and $\textrm{M}_{1}^{(i)}$.

A clear problem with this method, is that to perform SVD on high rank tensors, one already needs to be able to operate on the flattened tensor. If there are many agents or possible states, this matrix will be too big to be stored. In this case, one must resort to estimating the SVD decomposition. Consider the $\mathbf{M}_{t}^{J}$ tensor, in the case where there were many agents, and thus it was too big to be operated on. To decompose this, one would begin by creating the decomposed tensors $\mathbf{M}_{t}^{(i)}$ with initially random values. When doing so they would set the size of the bond dimension so that the tensors were small enough to be operated on by their machine. To fill in the values of the tensors, an approach can be used that is similar to the model learning part of section \ref{planning}. As long as it is accessible, one can sample many trajectories from a true environment. Then using the state transitions that occurred during these trajectories, they can update the $\mathbf{M}_{t}^{(i)}$ tensors using gradient descent. In doing this, one is minimising the Kullback-Leibler divergence between the empirical distribution (from the trajectories) and the model distribution. Implementing this goes beyond the scope of this project, and it would be a good piece of additional research.

\section{Policy Optimisation}
One final thing that must be discussed in this multi-agent case is policy optimisation. Luckily, one may follow a similar method to what was done in the single agent case, with one important difference. As before, to optimise the policy for an agent at a given timestep, one must first contract the rest of the network, being sure to do this in such a way that unnecessarily high rank tensors do not form. The important difference is that there now exists a policy tensor for each agent at each timestep, and the order in which one optimises these may make a significant difference to whether the true optimal policy is found. For example, the reward structure may be designed so that agent 1's rewards depends on agent 2, but agent 2's rewards do not depend on agent 1. In this case, in each timestep it would be intuitive to optimise the policy of agent 2 first, so that agent 1's policy was not being updated with respect to a version of agent 2's policy that was going to change anyway. 

\begin{figure}[h]
    \centering
    \includegraphics[width=13cm]{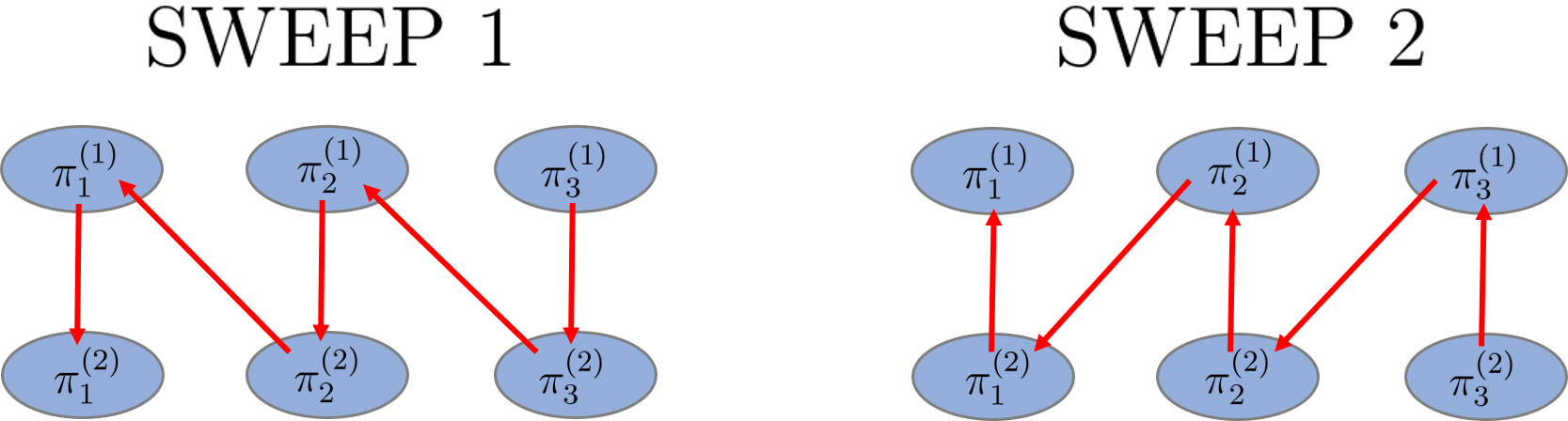}
    \caption{A figure showing the order at which to optimise the policy tensors. The red arrows show the next policy to optimise after the current one. SWEEP 1 begins by optimising $\bm{\pi}_{3}^{(1)}$, and SWEEP 2 begins with $\bm{\pi}_{3}^{(2)}$. }
    \label{policyorder}
\end{figure}

To avoid this, it may be worth sweeping through the policies twice, first optimising agents policies in numerical order (agent 1, agent 2...), and then the second time optimising agents policies in reverse numerical order (..., agent 2, agent 1). For the example in figure \ref{decomposemarlfinal}, this would involve optimising the policy tensors in the order shown in figure \ref{policyorder}. Of course, as the user sets up the rewards structure, the order of policies to optimise may be obvious.  

\chapter{MARL Example}

\ifpdf
    \graphicspath{{Chapter6/Figs/Raster/}{Chapter6/Figs/PDF/}{Chapter6/Figs/}}
\else
    \graphicspath{{Chapter6/Figs/Vector/}{Chapter6/Figs/}}
\fi
This chapter displays a simulation that is very similar to that of chapter 4, but in this case it involves a second agent. To perform the simulation, the joint representation was used rather than the decomposed representation, as only two agents were being used so it was not overly computationally exhaustive. Each agent received the same reward pattern as the single agent in chapter 4, except there were some further rewards that created the dependence between the agents. As before, results were collected for both a deterministic environment, and for an environment with normally distributed noise.

\section{Rewards}
Using the notation for the state of agents 1 and 2 being $S_{t}^{(1)}$ and $S_{t}^{(2)}$ respectively, the objective for this experiment was the following:

\begin{equation}
    \textrm{OBJECTIVE} = \:\:\: 
    \begin{cases}
    S_{t}^{(1)}>S_{t}^{(2)}\geq 0, & \text{if}\ t <T \\
    S_{t}^{(1)},S_{t}^{(2)} = 0, & \text{if}\ t = T
    \end{cases}
    \label{objectivemarl}
\end{equation}

In other words, the objective is composed of multiple parts. Firstly agent 1 must stay above agent 2 for all $t<T$, while both agents must be at or above $S = 0$. At time $t=T$, both agents must be at exactly $S=0$. One further addition to this, is that the objective that agent 1 must stay above agent 2 is given more importance than the agents being at or above $S=0$. To implement this, a larger penalty is awarded if agent 1 is not above agent 2, than if either of the agents are below 0. 

To encode this objective, each agent had its own reward structure, which are shown below.  Essentially, if agent 2 is not above agent 1 at any timestep $t<T$, both agents get a reward of -2 (additionally to the rewards of -1 for being below 0). Finally, both agents recieve a reward of 1 if they are at $S=0$ at $t=T$ and a reward of -10 if they are not. \newline \newline

\begin{center}
     $\textrm{IF}\:\: t<T:$
\end{center}
\begin{align*}
    r_{t}^{(i)} =
    \begin{cases}
    0, & \text{if}\ S_{t}^{(i)} \geq 0 \ \text{and}\ S_{t}^{(1)}<S_{t}^{(2)}\\
    -1, & \text{if}\ S_{t}^{(i)} < 0 \ \text{and}\ S_{t}^{(1)}<S_{t}^{(2)}\\
    -2, & \text{if}\ S_{t}^{(i)} \geq 0 \ \text{and}\ S_{t}^{(1)}\geq S_{t}^{(2)}\\
    -3, & \text{if}\ S_{t}^{(i)} < 0 \ \text{and}\ S_{t}^{(1)}\geq S_{t}^{(2)}\\
    \end{cases}\\
\end{align*}
\begin{center}
    $\textrm{ELSE IF}\:\: t=T:$   
\end{center}
\begin{align*}
    r_{t}^{(i)} =
    \begin{cases}
    1, & \text{if}\ S_{t}^{(i)} = 0\\
    -10, & \text{otherwise}
    \end{cases}\\
\end{align*}

\section{Results - Sample Trajectories}
The agent's joint policy was initialised as random ($P(a^{(1)}_{t},a^{(2)}_{t}|s^{(1)}_{t},s^{(2)}_{t}) = \frac{1}{(N_{A})^{2}} \, \forall \, a,s,t$). The initial expected return was calculated by contracting the network, and 100 trajectories were generated. The policy was then optimised before the expected return was calculated again, and 100 more trajectories were sampled. All simulations were done using $T=6$, due to time constraints.

\subsubsection{Deterministic Environment}
\begin{figure}[H]
     \centering
     \begin{subfigure}[b]{0.49\textwidth}
         \centering
         \includegraphics[width=\textwidth]{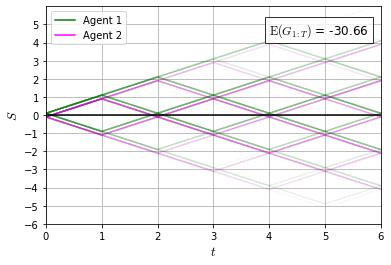}
         \caption{Before Policy Optimisation.}
         \label{fig:y equals x}
     \end{subfigure}
     \begin{subfigure}[b]{0.49\textwidth}
         \centering
         \includegraphics[width=\textwidth]{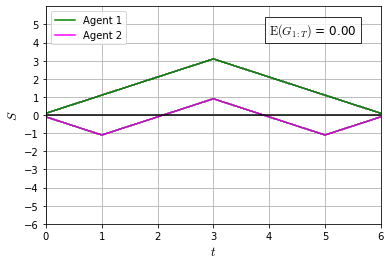}
         \caption{After Policy Optimisation.}
         \label{fig:deterministic optimise}
     \end{subfigure}
     \hfill
        \caption{Showing 100 trajectories for each agent, that were generated before and after policy optimisation. Agent 1's trajectories are marked in green and agent 2's are in magenta, as well as the lines being plotted slightly offset from each other for clarity. a) The random policy generates trajectories where there is an equal chance of going up or down for both agents. b) There is only one possible pair of trajectories that obeys the objective. The optimised policy exclusively generates this pair. For agent 2 there is more punishment for not being below agent 1 than there is for spending a timestep below 0. Therefore it chooses to go down in timesteps 0 and 4.}
        \label{fig:deterministicmarl}
\end{figure}

Figure \ref{fig:deterministicmarl} displays the results of the 100 trajectories for the deterministic environment before and after policy optimisation. Before optimisation there is equal chance of going up or down for each agent. Therefore the distribution of both agent's positions remains fairly similar throughout the timesteps. In the SARL example, once the policy was optimised there were many trajectories that could be generated, all of which satisfied the objective. In this MARL example, after optimisation we see that the same pair of trajectories is generated for all 100 samples, due to the fact that this is the only one that perfectly obeys the objective. Agent 2 learns that when faced with the choice between going below 0, or being in the same state as agent 1, to always go below 0. This is because the rewards were set up to give a smaller punishment for doing so.  

\subsubsection{Environment with Normal Random Noise}
\begin{figure}[H]
     \centering
     \begin{subfigure}[b]{0.49\textwidth}
         \centering
         \includegraphics[width=\textwidth]{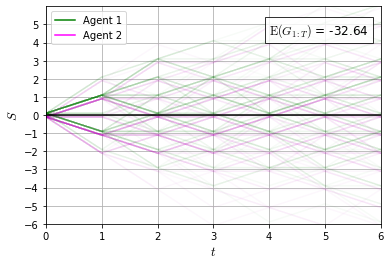}
         \caption{Before Policy Optimisation.}
         \label{fig:y equals x}
     \end{subfigure}
     \begin{subfigure}[b]{0.49\textwidth}
         \centering
         \includegraphics[width=\textwidth]{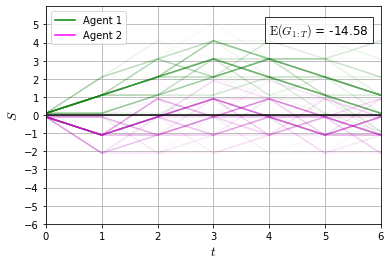}
         \caption{After Policy Optimisation.}
         \label{fig:three sin x}
     \end{subfigure}
     \hfill
        \caption{An analogous diagram to figure \ref{fig:deterministicmarl}, but with the presence of random noise. a) The trajectories are more diffuse due to the possibility of moving two states at once. b) Agent 1 never goes below $S=0$. Agent 2 then attempts to remain above $S=0$, but not if it means going above agent 1. Due to the noise, there is no guarantee that the agents will end at $S=0$.}
        \label{fig:normalmarl}
\end{figure}

Normally distributed random noise was added once more, again with $\sigma = 1$. This gives either agent a possibility of jumping 0 or 2 states in one timestep. The results of generating 100 trajectories before and after policy optimisation are shown in figure \ref{fig:normalmarl}. Before policy optimisation the expected return is lower than in the deterministic case, for the same reasons in the SARL example. Due to the possibility of moving 0 or 2 states, there are more states that it is possible to land on at $t=T$. Therefore there is more chance of obtaining the large punishment of -10. After policy optimisation we see trajectories with an average of roughly the same as seen in figure \ref{fig:deterministic optimise}, but with deviations due to the noise. It is worth noting that there exists pairs of trajectories that will now score a better return than $G = 0$. For example, if agent 2 takes the action to go down at time $t=0$, but the noise causes its state to remain the same, it will not receive the punishment for being beneath $s = 0$. With this in mind the actual best possible pairs of trajectories receive a return of $G = +2$. Due to the noise, there is no way of reliably accessing these pairs however, and the overall effect that the noise causes on the expected return is detrimental.

\section{Example - SVD}
In the examples in this section, $T = 6$, which meant that $N_{S} = 13$. The number of possible actions was $N_{A} = 2$ and the number of rewards was $N_{R} = 6$. In this case the number of elements in the rank 8 $\mathbf{M}_{t}^{J}$ tensor is equal to $13^{4} \times 2^{2} \times 6^{2} \approx 4\times 10^{6} $. The machine that was used to perform these simulations was already near the peak of its performance at this point, and had $T$ been larger than 6, the device would have not had enough memory to store the $\mathbf{M}_{t}^{J}$ tensor. This situation provides a good opportunity to implement SVD, and to see if the information in $\mathbf{M}_{t}^{J}$ can be compressed into two tensors with fewer elements. To do this, the following process was repeated many times for both the deterministic and noisy environments, varying the Schmidt rank each time. 

\begin{figure}[H]
    \centering
    \includegraphics[width=10cm]{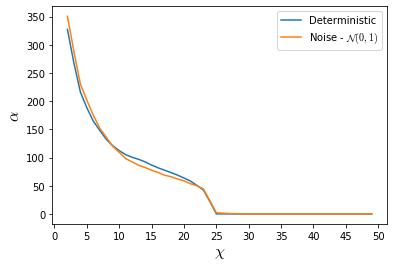}
    \caption{Figure showing $\alpha$ when $_{\textrm{SVD}}\mathbf{M}_{t}^{J}$ was created using a certain Schmidt rank, $\chi$. $\alpha$ was reduced to 0 at a Schmidt rank of 25 and 26 for the deterministic and noisy cases respectively.}
    \label{fig:svderror}
\end{figure}

First the $\mathbf{M}_{t}^{J}$ tensor was flattened into a $2028\times 2028$ matrix, with indices according to the variables for each agent (see figure \ref{fig:svdm}). SVD was then used to decompose the flattened $\mathbf{M}_{t}^{J}$ tensor, into two matrices $\mathbf{M}_{t}^{(1)}$ and $\mathbf{M}_{t}^{(2)}$. The size of these tensors depended on the Schmidt rank. These two tensors were then contracted together by the shared index, to reform a tensor analogous to the flattened $\mathbf{M}_{t}^{J}$ tensor. Such tensor is then unrolled back into a rank 8 tensor, which one can name $_{\textrm{SVD}}\mathbf{M}_{t}^{J}$. Finally, the absolute element wise difference between $_{\textrm{SVD}}\mathbf{M}_{t}^{J}$ and $\mathbf{M}_{t}^{J}$ is found, according to the following formula:

\begin{equation}
    \alpha = \sum|(_{\textrm{SVD}}\mathbf{M}_{t}^{J} - \mathbf{M}_{t}^{J})|
\end{equation}
Where in this case, the $\sum$ symbol represents the sum over all elements. The resulting error was then plotted against the Schmidt rank, seen in figure \ref{fig:svderror}. 

For the deterministic $\mathbf{M}_{t}^{J}$, one sees that $\alpha$ reaches 0 for a value of $\chi = 25$. The result of this is that when flattened, $\mathbf{M}_{t}^{(1)}$ and $\mathbf{M}_{t}^{(2)}$ are each $2028\times 25$ matrices, that when contracted, have the ability to perfectly represent the flattened $\mathbf{M}_{t}^{J}$ $2028\times 2028$ matrix. This is hugely significant as the combined total number of elements in $\mathbf{M}_{t}^{(1)}$ and $\mathbf{M}_{t}^{(2)}$ is 101400, which is approximately 2.5\% of the number of elements in $\mathbf{M}_{t}^{J}$. It is also interesting to see the similarity between the deterministic and noisy cases. Intuitively, one may think that it would take a lower value of $\chi$ to represent the deterministic $\mathbf{M}_{t}^{J}$, as the tensor contains only the digits 1 and 0. However, we actually see that it only takes an increase of $\chi$ by 1 for $\alpha$ to go to 0 in the noisy case.
\chapter{Conclusion}  

\ifpdf
    \graphicspath{{Conclusion/Figs/Raster/}{Conclusion/Figs/PDF/}{Conclusion/Figs/}}
\else
    \graphicspath{{Conclusion/Figs/Vector/}{Conclusion/Figs/}}
\fi

In this report, a general tensor network representation of the expected return of a multi-agent reinforcement learning task has been found, under the assumption that the task can be described as a finite Markov decision process. This structure was then applied to a two-agent random walker example, in which it was proven that the policy could be correctly optimised. Finally, the power of singular value decomposition was shown, where it was used to reduce the number of elements in a tensor by 97.5\% without experiencing any loss of information.

\section{Further Work}
If one were to continue the work that has been described in this report, a natural next step would be to create a TN for many agents, and implementing the model learning ideas that were introduced at the end of section \ref{decompositionsection}. It would be interesting to observe the degree to which the tensors could be compressed, whilst still outputting realistic results. Other further work could involve testing tensor network's speed against existing distribution models, and potentially broadening the ideas to competitive MARL.

\begin{spacing}{0.9}

\cleardoublepage

\end{spacing}

\printthesisindex

\end{document}